\documentclass[utf8]{include/frontiersSCNS} % for Science, Engineering and Humanities and Social Sciences articles
\usepackage{url,hyperref,lineno,microtype,subcaption}
\usepackage[onehalfspacing]{setspace}
\usepackage{amsmath}
\usepackage{bm}
\pdfoutput=1 % for arxiv

\newcommand{\figref}[1]{{Figure \ref{#1}}}

\newcommand{\equref}[1]{{Eq. (\ref{#1})}}

\newcommand{\secref}[1]{Section \ref{#1}}

\newcommand{\switchlanguage}[2]{%
  \ifx\paperlanguage\empty%
  #1%
  \else%
  #2%
  \fi%
}

\def\paperlanguage{}

\def\keyFont{\fontsize{8}{11}\helveticabold }
\def\firstAuthorLast{Kawaharazuka {et~al.}} %use et al only if is more than 1 author
\def\Authors{Kento Kawaharazuka\,$^{1,*}$, Akihiro Miki\,$^{1}$, Masahiro Bando\,$^{1}$, Kei Okada\,$^{1}$, and Masayuki Inaba\,$^{1}$}
\def\Address{$^{1}$ JSK Robotics Laboratory, Department of Mechano-Informatics, Graduate School of Information Science and Technology, The University of Tokyo, Bunkyo-ku, Tokyo, JAPAN}
\def\corrAuthor{Kento Kawaharazuka}

\def\corrEmail{kawaharazuka@jsk.imi.i.u-tokyo.ac.jp}

\begin{document}
\onecolumn
\firstpage{1}

\title[Dynamic Cloth Manipulation]{Dynamic Cloth Manipulation Considering Variable Stiffness and Material Change Using Deep Predictive Model with Parametric Bias}

\author[\firstAuthorLast ]{\Authors} %This field will be automatically populated
\address{} %This field will be automatically populated
\correspondance{} %This field will be automatically populated

\extraAuth{}% If there are more than 1 corresponding author, comment this line and uncomment the next one.
%\extraAuth{corresponding Author2 \\ Laboratory X2, Institute X2, Department X2, Organization X2, Street X2, City X2 , State XX2 (only USA, Canada and Australia), Zip Code2, X2 Country X2, email2@uni2.edu}

\maketitle

\begin{abstract}
\switchlanguage%
{%
  Dynamic manipulation of flexible objects such as fabric, which is difficult to modelize, is one of the major challenges in robotics.
  With the development of deep learning, we are beginning to see results in simulations and in some actual robots, but there are still many problems that have not yet been tackled.
  Humans can move their arms at high speed using their flexible bodies skillfully, and even when the material to be manipulated changes, they can manipulate the material after moving it several times and understanding its characteristics.
  Therefore, in this research, we focus on the following two points: (1) body control using a variable stiffness mechanism for more dynamic manipulation, and (2) response to changes in the material of the manipulated object using parametric bias.
  By incorporating these two approaches into a deep predictive model, we show through simulation and actual robot experiments that Musashi-W, a musculoskeletal humanoid with variable stiffness mechanism, can dynamically manipulate cloth while detecting changes in the physical properties of the manipulated object.
}%
{%
  モデル化が難しい布のような柔軟物の動的マニピュレーションはロボットにおける大きな課題の一つである.
  深層学習の発展により, シミュレーションや一部の実機において成果が出始めているが, まだ取り組まれていない課題も多い.
  特に人間は, その柔軟な身体を巧みに使って手先を高速に動かし, 操作対象の素材が変化しても, 少し動かして特性を理解したうえでマニピュレーションを行うことができる.
  そこで本研究では, (1)より動的なマニピュレーションのための可変剛性機構を用いた身体制御, (2) Parametric Biasを用いた操作対象物体の素材変化への対応, という2点に焦点を当てる.
  この2つの取り組みを深層予測モデルに組み込むことで, 可変剛性制御可能な筋骨格ヒューマノイドMusashi-Wが, 操作対象の物性変化を検知しつつ動的に布をマニピュレーション可能となることを, 実機実験により示す.
}%
\tiny
 \keyFont{\section{Keywords:} deep learning, predictive model, cloth manipulation, variable stiffness, parametric bias, musculoskeketal humanoid} %All article types: you may provide up to 8 keywords; at least 5 are mandatory.
\end{abstract}

\section{Introduction}\label{sec:introduction}
\switchlanguage%
{%
  Manipulation of flexible objects such as fabric, which is difficult to modelize, is one of the major challenges in robotics.
  There are two main types of manipulation of flexible objects: static manipulation and dynamic manipulation.
  For each type, various model-based and learning-based methods have been developed.
  Static manipulation has been tackled for a long time, and various model-based methods exist \citep{inaba1987rope, saha2007deformable, elbrechter2012folding}.
  Learning-based methods have also been actively pursued in recent years and have been successfully applied to actual robots \citep{lee2015learning, tanaka2018emd}.
  On the other hand, there are not so many examples of dynamic manipulation, which is more difficult compared to static manipulation.
  As for model-based methods, the work of Yamakawa et al. on cloth folding and knotting is well known \citep{yamakawa2010knotting, yamakawa2011folding}.
  Learning-based methods include \citep{kawaharazuka2019dynamic}, which uses deep predictive models, and \citep{jangir2020cloth}, which uses reinforcement learning.
  Also, there are many examples where only static manipulation is performed even though dynamic models are used \citep{hoque2020cloth, ebert2018foresight}.
  There is an example where the primitive of the dynamic motion is generated manually and only the grasping point is trained \citep{ha2021flingbot}.
  % Note that \citep{kawaharazuka2019dynamic} uses an actual robot with one degree of freedom, while \citep{jangir2020cloth} uses a robot with multiple degrees of freedom, but only in simulation.
  % We focus on the dynamic manipulation in this study.
  % In order to deal with multiple elements that are difficult to modelize as described subsequently, we develop a learning-based method through trial and error.
  % In order to apply the method to the actual robot, we develop a dynamic cloth manipulation method with a deep predictive model instead of reinforcement learning, which requires a large number of trials.
}%
{%
  モデル化の難しい布のような柔軟物のマニピュレーションはロボットにおける大きな課題の一つであり, 現在盛んに研究されている.
  これら柔軟物体マニピュレーションには大まかに, 静的マニピュレーションと動的マニピュレーションが存在する.
  また, それぞれについて, model-basedな手法・learning-basedな手法が様々に開発されてきた.
  静的なマニピュレーションは古くから取り組まれており, 様々なmodel-basedな手法が存在する\citep{inaba1987rope, saha2007deformable, elbrechter2012folding}.
  learning-basedな手法も近年盛んに取り組まれており, 実機への応用にも成功している\citep{lee2015learning, tanaka2018emd}.
  一方, 静的マニピュレーションに比べてより難しい動的なマニピュレーションの例はそこまで多くはない.
  model-basedな手法としては, 山川らのcloth foldingやknottingの研究がよく知られている\citep{yamakawa2010knotting, yamakawa2011folding}.
  機械学習ベースの手法としては, 深層予測モデルを使ったもの\citep{kawaharazuka2019dynamic}, 強化学習を使ったもの\citep{jangir2020cloth}がある.
  また, 動的なモデルを使っていても, 静的なマニピュレーションしか行っていない例は多い\citep{hoque2020cloth, ebert2018foresight}.
  動的動作のprimitiveは手動で生成した上で, 把持点のみ学習を行った例もある\citep{ha2021flingbot}.
  % なお, \citep{kawaharazuka2019dynamic}は最大2自由度の実機のみを使っており, \citep{jangir2020cloth}は多自由度ロボットを使っているものの, シミュレーションのみでの実行である.
  % 後に述べるようなモデル化の容易ではない要素を複数扱うため, 試行錯誤型のlearning-basedな手法を開発する.
  % その中でも, 実機への適用を行うため, 多数の試行が必要な強化学習ではなく, 深層予測モデルによる動的布操作を行う.
}%

\begin{figure}[t]
  \centering
  \includegraphics[width=0.6\columnwidth]{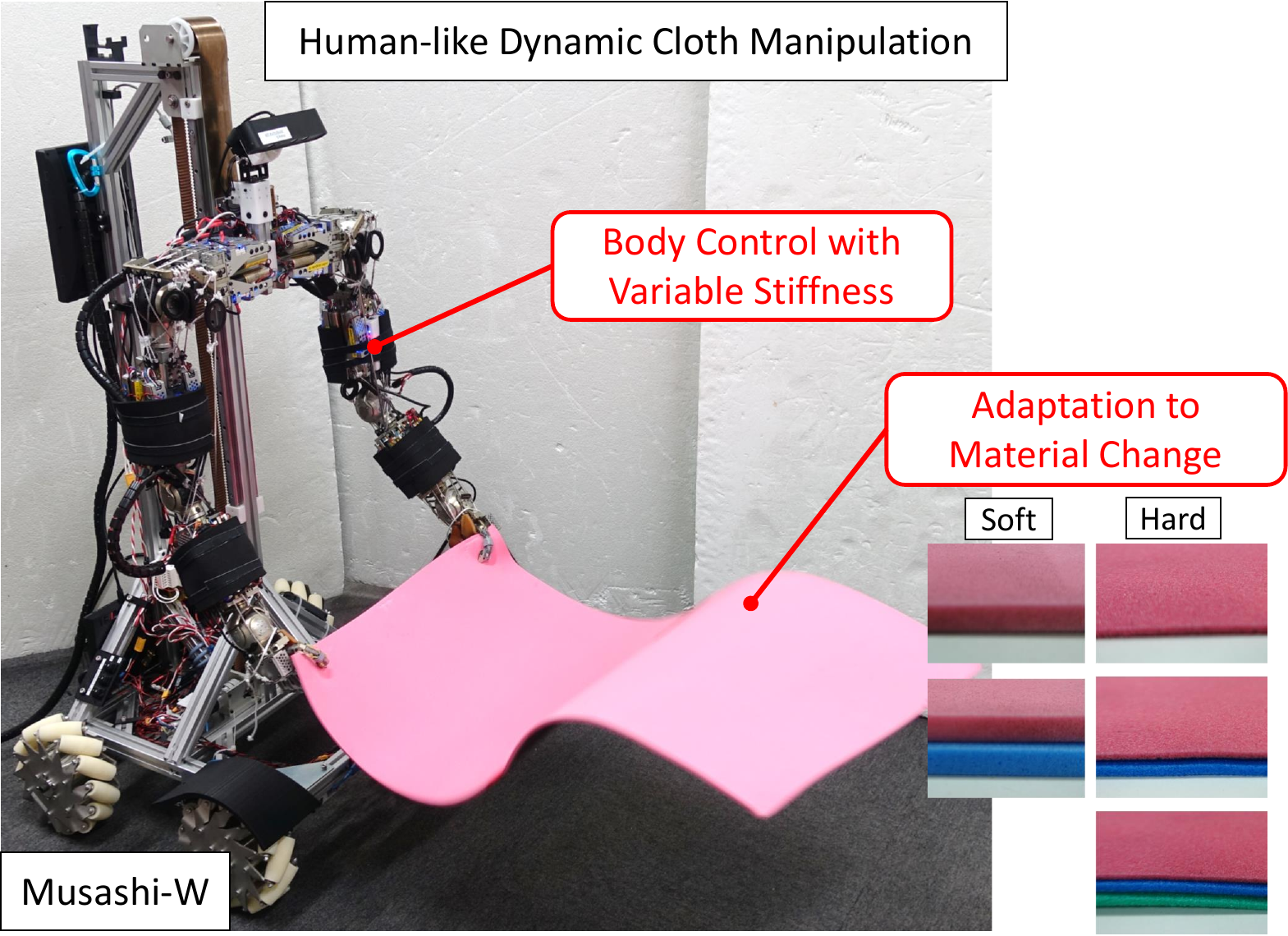}
  \caption{Dynamic cloth manipulation by the musculoskeletal humanoid Musashi-W considering body control with variable stiffness and adaptation to material change.}
  \label{figure:concept}
\end{figure}

\switchlanguage%
{%
  In this study, we handle a dynamic manipulation such as the spreading of a bed sheet or a picnic sheet.
  For the dynamic manipulation, there are several points lacking in human-like adaptive dynamic cloth manipulation that have not been addressed in the introduced previous studies.
  The previous studies have not been able to utilize the flexible bodies of robots to move their arms at high speed as humans do, and to perform manipulation based on an understanding of the characteristics of a material from a few trials, even if the material to be manipulated changes.
  Therefore, we focus on two points: (1) body control for more dynamic manipulation, and (2) adaptation to changes in the material of the manipulated object (\figref{figure:concept}).
  In (1), we aim at cloth manipulation by a robot with variable stiffness mechanism, which is flexible like a human and can manipulate its flexibility at will.
  In this study, we perform manipulation by Musashi-W (MusashiDarm \citep{kawaharazuka2019musashi} with wheeled base), a musculoskeletal humanoid with variable stiffness mechanism using redundant muscles and nonlinear elastic elements.
  We consider how the dynamic cloth manipulation is changed by using the additional stiffness value as the control command of the body.
  There have been examples of dynamic pitching behavior by changing hardware stiffness in the framework of model-based optimal control \citep{braun2012variable}, and several static tasks by changing software impedance using reinforcement learning \citep{martin2019variable}.
  On the other hand, there is no example focusing on changes in the hardware stiffness in a dynamic handling task of difficult-to-modelize objects that requires learning-based control, such as in this study.
  In (2), we aim at adapting to changes in the material of the manipulated object using parametric bias \citep{tani2002parametric}.
  The parametric bias is an additional bias term in neural networks, which has been mainly used for imitation learning to extract multiple attractor dynamics from various motion data \citep{ogata2005extracting, kawaharazuka2021imitation}.
  In this study, we use it to embed information about the material and physical properties of the cloth.
  When the robot holds a new cloth, the physical properties can be identified by manipulating the cloth several times, and dynamic cloth manipulation can be accurately performed.
  Parametric bias can be applied not only to simulations but also to actual robots where it is difficult to identify the parameters of cloth materials, since it implicitly self-organizes from differences in dynamics of each data.
  In this study, we construct a deep predictive model that incorporates (1) and (2), and demonstrate the effectiveness by performing a more human-like dynamic cloth manipulation.

  The contribution of this study is as follows.
  \begin{itemize}
      \item Examination of the effect of adding the body stiffness value to the control command
      \item Adaptation to changes in cloth material using parametric bias
      \item Human-like adaptive dynamic cloth manipulation by learning a deep predictive model considering variable stiffness and material change
  \end{itemize}

  % This study is organized as follows.
  In \secref{sec:proposed-method}, we describe the structure of the deep predictive model, the details of variable stiffness control, the estimation of material parameters, and the body control for dynamic cloth manipulation.
  In \secref{sec:experiments}, we discuss online learning of material parameters and changes in dynamic cloth manipulation due to variable stiffness and cloth material changes in simulation and the actual robot.
  In addition, a table setting experiment including dynamic cloth manipulation is conducted.
  The results of the experiments are discussed in \secref{sec:discussion}, and the conclusion is given in \secref{sec:conclusion}.
}%
{%
  本研究では, まだ例の少ない, ベッドシーツやレジャーシートを大きく広げるような, 動的マニピュレーションについて扱う.
  一方で, これまで紹介した先行研究において取り組まれていない, 人間のような適応的な動的布操作に足りない点がいくつかある.
  これまでの研究は, 人間のように, その柔軟な身体を使って手先を高速に動かし, 操作対象の素材が変化しても少しの試行からその素材の特性を理解したうえでマニピュレーションを行うことができていない.
  そこで本研究では, (1)より動的なマニピュレーションのための身体制御, (2)操作対象物体の素材変化への対応, という2点に新しく焦点を当てる(\figref{figure:concept}).
  (1)は, 人間のように柔軟かつその柔軟さを自在に操ることのできる可変剛性制御が可能なロボットによるマニピュレーションを目指す.
  本研究では冗長な筋肉と非線形弾性要素により可変剛性制御可能な筋骨格ヒューマノイドMusashi-W (MusashiDarm \citep{kawaharazuka2019musashi}に車輪が付いたロボット)によりマニピュレーションを行う.
  このとき, 身体の制御入力として, 追加で身体剛性値を用いることで, どのように動的布操作が変化するかを考察する.
  これまで, モデルベースな手法によりハードウェア剛性値を最適制御の枠組みで変化させ動的な投球動作を行った例\citep{braun2012variable}や, 静的なタスクにおいてソフトウェアのインピーダンス変化を獲得する学習手法が存在する\citep{martin2019variable}.
  一方, 本研究のようなモデル化困難な物体を動的に扱う学習型制御を要するタスクにおいて, 身体のハードウェア剛性変化に着目した例はない.
  (2)は, Parametric Bias \citep{tani2002parametric}を用いた操作対象の素材変化への適応を目指す.
  Parametric Biasはニューラルネットワークにおける追加のバイアス項であり, 様々な動作データについて複数のattractor dynamicsを抽出することを目的として, 主に模倣学習に利用されてきた\citep{ogata2005extracting, kawaharazuka2021imitation}.
  本研究ではこれを, 布の素材, 物理的特性情報を埋め込むために用いる.
  新しい布を持ったとき, これを少し操作することで物理的特性を同定し, これによって正確に動的布操作を行う.
  Parametric Biasはデータごとのダイナミクスの違いから, 特にパラメータ等を与えずとも暗黙的に自己組織化するため, シミュレーションだけでなく, 布素材パラメータの同定が難しい実機においても適用可能である.
  本研究では, この(1)と(2)を組み込んだ深層予測モデルを構築し, より人間らしい動的布操作を行うことでその有効性を示す.

  本研究のコントリビューションは以下である.
  \begin{itemize}
      \item 制御入力への身体剛性値の追加とその効果の考察
      \item Parametric Biasによる布の素材変化への適応
      \item 可変剛性と素材変化を考慮した深層予測モデル学習よる人間らしい適応的な動的布操作
  \end{itemize}

  本研究は以下のような構成となっている.
  \secref{sec:proposed-method}では深層予測モデルの構造, 可変剛性制御の導入, 素材パラメータの推定, 動的布状態制御について述べる.
  \secref{sec:experiments}では, シミュレーションと実機において, 素材パラメータのオンライン学習, 可変剛性・布素材変化による動的布操作の変化について考察する.
  また, 動的布操作を含む一連のテーブルセッティング実験も行う.
  \secref{sec:discussion}では実験結果について議論し, \secref{sec:conclusion}で結論を述べる.
}%

\begin{figure}[t]
  \centering
  \includegraphics[width=1.0\columnwidth]{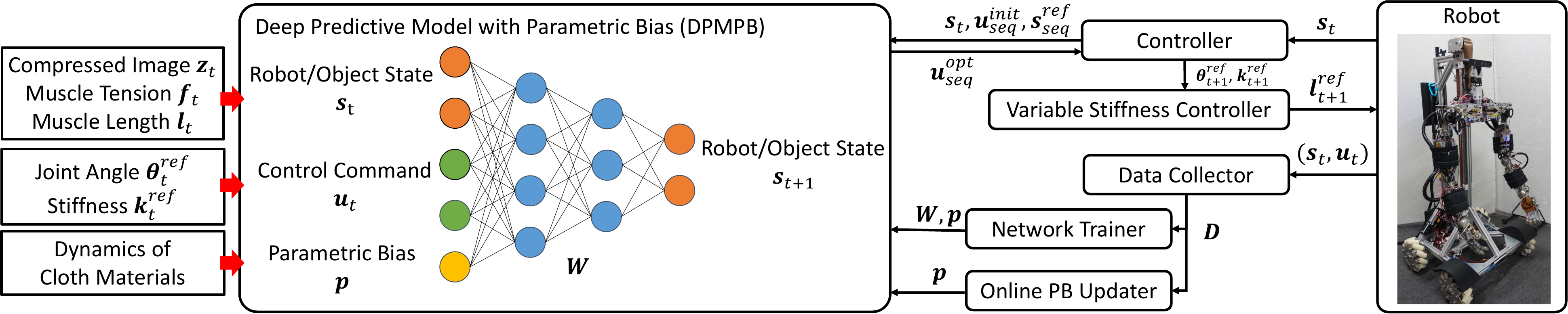}
  \caption{The overview of our system: deep predictive model with parametric bias (DPMPB), controller using DPMPB for dynamic cloth manipulation, variable stiffness controller for musculoskeletal humanoids, data collector for DPMPB, and online updater of parametric bias (PB).}
  \label{figure:whole-system}
\end{figure}

\section{Dynamic Cloth Manipulation Considering Variable Stiffness and Material Change} \label{sec:proposed-method}
\switchlanguage%
{%
  We call the network used in this study Dynamic Predictive Model with Parametric Bias (DPMPB).
  The entire system is shown in \figref{figure:whole-system}.
}%
{%
  本研究で用いるネットワークを, Deep Predctive Model with Parametric Bias (DPMPB)と呼ぶ.
  全体システムを\figref{figure:whole-system}に示す.
}%

\subsection{Network Structure of DPMPB} \label{subsec:network-structure}
\switchlanguage%
{%
  DPMPB can be expressed by the following equation.
  \begin{linenomath}
    \begin{align}
      \bm{s}_{t+1} = \bm{h}_{dpmpb}(\bm{s}_{t}, \bm{u}, \bm{p}) \label{eq:dpmpb}
    \end{align}
  \end{linenomath}
  where $t$ is the current time step, $\bm{s}$ is the state of the manipulated object and robot, $\bm{u}$ is the control command to the robot body, $\bm{p}$ is the parametric bias (PB), and $\bm{h}_{dpmpb}$ is the function representing the time series change in the state of the manipulated object and robot due to the control command.
  We use the state of the cloth and the state of the robot for $\bm{s}$ in this study of cloth manipulation.
  For the cloth state, we use $\bm{z}_{t}$, which is the compressed value of the current image $I_{t}$ by using AutoEncoder \citep{hinton2006reducing}.
  Although the robot state should be different for each robot, we use $\bm{f}_{t}$ and $\bm{l}_{t}$ for the musculoskeletal humanoid used in this study ($\bm{f}_{t}$ and $\bm{l}_{t}$ represent the current muscle tension and muscle length).
  Thus, $\bm{s}^{T}_{t} = \begin{pmatrix}\bm{z}^{T}_{t} & \bm{f}^{T}_{t} & \bm{l}^{T}_{t}\end{pmatrix}$.
  Also, we set $\bm{u}^{T}=\begin{pmatrix}\bm{\theta}^{ref, T} & \bm{k}^{ref, T}\end{pmatrix}$ ($\bm{\theta}^{ref}$ represents the target joint angle and $\bm{k}^{ref}$ represents the target body stiffness value).
  The details of the target joint angle and body stiffness are described in \secref{subsec:variable-stiffness}.
  The parametric bias is a value that can embed implicit differences in dynamics, which are common for the same material and different for different materials.
  By collecting data using various cloth materials, information on the dynamics of the material of the manipulated object is embedded in $\bm{p}$.
  Note that each sensor value is used as input to the network after normalization using all the obtained data.

  In this study, the DPMPB consists of 10 layers: 4 fully-connected layers, 2 LSTM layers \citep{hochreiter1997lstm}, and 4 fully-connected layers, in order.
  The number of units is set to \{$N_u+N_s+N_p$, 300, 100, 30, 30 (number of units in LSTM), 30 (number of units in LSTM), 30, 100, 300, $N_s$\} (where $N_{\{u, s, p\}}$ is the number of dimensions of $\{\bm{u}, \bm{s}, \bm{p}\}$).
  The activation function is hyperbolic tangent and the update rule is Adam \citep{kingma2015adam}.
  Regarding the AutoEncoder, the image is compressed by applying convolutional layers with kernel size 3 and stride 2 five times to a $128\times96$ binary image (the cloth part is extracted in color), then reducing the dimensionality to 256 and 3 units, in order, using the fully-connected layers, and finally restoring the image by the fully-connected layers and the deconvolutional layers.
  For all layers exempting the last layer, the batch normalization \citep{ioffe2015batchnorm} is applied, and the activation function is ReLU \citep{nair2010relu}.
  The dimension of $\bm{p}$ is set to 2 in this study, which should be sufficiently smaller than the number of cloth materials used for training.
  The execution period of \equref{eq:dpmpb} is set to 5 Hz.
}%
{%
  本研究で用いるDPMPBは以下のように数式で表せる.
  \begin{linenomath}
    \begin{align}
      \bm{s}_{t+1} = \bm{h}_{dpmpb}(\bm{s}_{t}, \bm{u}, \bm{p}) \label{eq:dpmpb}
    \end{align}
  \end{linenomath}
  ここで, $t$は現在のタイムステップ, $\bm{s}$は操作物体やロボットの状態, $\bm{u}$はロボット身体への制御入力, $\bm{p}$はParametric Bias, $\bm{h}_{dpmpb}$はロボットと操作物体状態の制御入力による時系列変化を表す関数である.
  $\bm{s}$は, 本研究の布操作では, 操作物体である布の状態と, ロボットの状態を用いる.
  布の状態は, 得られた現在の画像$I_{t}$をAutoEncoder \citep{hinton2006reducing}で圧縮した$\bm{z}_{t}$を用いる.
  ロボットの状態は, ロボットごとに異なるべきであるが, 本研究で扱う筋骨格ヒューマノイドに合わせて, $\bm{f}_{t}$と$\bm{l}_{t}$とした($\bm{f}_{t}, \bm{l}_{t}$は現在の筋張力・筋長を表す).
  よって, $\bm{s}^{T}_{t} = \begin{pmatrix}\bm{z}^{T}_{t} & \bm{f}^{T}_{t} & \bm{l}^{T}_{t}\end{pmatrix}$である.
  また, 本研究では$\bm{u}^{T}=\begin{pmatrix}\bm{\theta}^{ref, T} & \bm{k}^{ref, T}\end{pmatrix}$とした(ここで, $\bm{\theta}^{ref}$は指令関節角度, $\bm{k}^{ref}$は指令身体剛性値を表す).
  指令関節角度, 身体剛性の詳細については\secref{subsec:variable-stiffness}に述べる.
  Parametric Biasは暗黙的なdynamicsの違いを埋め込むことのできる値であり, 同一の素材同士については共通, 異なる素材同士については互いに異なるような値である.
  様々な布の素材を使ってマニピュレーションを試すことで, $\bm{p}$には操作物体の素材のダイナミクスに関する情報が埋め込まれる.
  なお, それぞれのセンサ値は得られた全データを使って正規化してからネットワークに入力している.

  本研究においてDPMPBは10層とし, 順に4層の全結合層, 2層のLSTM層 \citep{hochreiter1997lstm}, 4層の全結合層からなる.
  ユニット数については, \{$N_u+N_s+N_p$, 300, 100, 30, 30 (LSTMのユニット数), 30 (LSTMのユニット数), 30, 100, 300, $N_s$\}とした(なお, $N_{\{u, s, p\}}$は$\{\bm{u}, \bm{s}, \bm{p}\}$の次元数とする).
  活性化関数はHyperbolic Tangent, 更新則はAdam \citep{kingma2015adam}とした.
  画像を圧縮するAutoEncoderの構造については, $128\times96$の2値画像(布部分を色抽出している)について, カーネルサイズが3, ストライドが2の畳み込み層を5回適用し, 全結合層で順にユニット数256, 3まで次元を削減したあと, 同様に全結合層・逆畳み込み層によって画像を復元していく形を取っている.
  最終層以外についてはBatch Normalization \citep{ioffe2015batchnorm}が適用され, 活性化関数は最終層以外についてはReLU \citep{nair2010relu}, 最終層はSigmoid, 更新則はAdam \citep{kingma2015adam}とした.
  $\bm{p}$の次元は, 使用した布の素材数よりも十分に小さい値とし, 本研究では2とした.
  また, \equref{eq:dpmpb}の実行周期は5Hzとする.
}%

\begin{figure}[t]
  \centering
  \includegraphics[width=0.6\columnwidth]{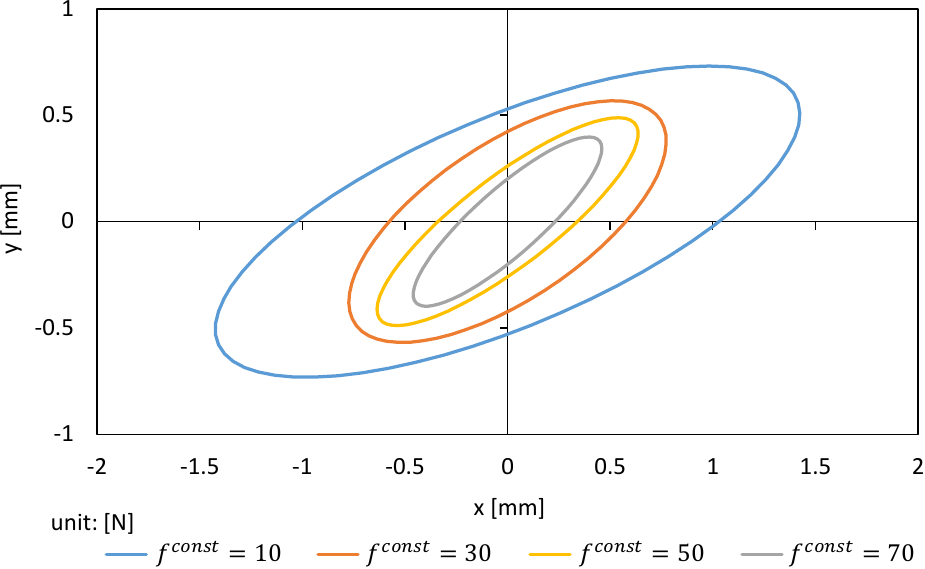}
  \caption{Operational stiffness ellipsoid when adding force of 1 N while changing $f^{const}=\{10, 30, 50, 70\}$ [N].}
  \label{figure:variable-stiffness}
\end{figure}

\subsection{Variable Stiffness Control for Musculoskeletal Humanoids} \label{subsec:variable-stiffness}
\switchlanguage%
{%
  Regarding the musculoskeletal humanoid with variable stiffness mechanism used in this study, we will describe the details of $\bm{\theta}^{ref}$ and $\bm{k}^{ref}$.
  In this study, we use the following relationship of joint angle $\bm{\theta}$, muscle tension $\bm{f}$, and muscle length $\bm{l}$ \citep{kawaharazuka2018bodyimage}.
  \begin{linenomath}
    \begin{align}
      \bm{l} = \bm{h}_{bodyimage}(\bm{\theta}, \bm{f}) \label{eq:bodyimage}
    \end{align}
  \end{linenomath}
  $\bm{h}_{bodyimage}$ is trained using the actual robot sensor information.
  When using this trained network for control, the target joint angle $\bm{\theta}^{ref}$ and the target muscle tension $\bm{f}^{ref}$ are determined and the corresponding target muscle length $\bm{l}^{ref}=\bm{h}_{bodyimage}(\bm{\theta}^{ref}, \bm{f}^{ref})$ is calculated.
  However, since this value is the muscle length to be measured, not the target muscle length, $\bm{l}^{send}$, which takes into account the muscle elongation in muscle stiffness control \citep{shirai2011stiffness}, is sent to the actual robot \citep{kawaharazuka2018bodyimage} in practice.

  Here, we consider how $\bm{f}^{ref}$ is given.
  Normally, $\bm{f}^{ref}$ should be the value required to realize $\bm{\theta}^{ref}$.
  On the other hand, in \citep{kawaharazuka2018bodyimage}, a constant value $f^{const}$ is first given to $\bm{f}^{ref}$ for all muscles to achieve $\bm{\theta}^{ref}$ to a certain degree, and then the current muscle tension $\bm{f}$ is given as $\bm{f}^{ref}$ to achieve $\bm{\theta}^{ref} $ more accurately.
  The body stiffness can be changed to some extent depending on the value of $f^{const}$ sent at the beginning.
  In addition, since the relationship between $\bm{s}$ and $\bm{u}$ is eventually acquired by learning in this study, it is not necessary to realize $\bm{\theta}^{ref}$ precisely.
  Therefore, we use $f^{const}$ as the body stiffness value $\bm{k}^{ref}$ included in $\bm{u}$ and operate the robot with $\bm{l}^{ref}=\bm{h}_{bodyimage}(\bm{\theta}^{ref}, \bm{f}^{const})$.
  By acquiring training data while changing $f^{const}$, we can perform dynamic cloth manipulation considering the body stiffness value.

  In the following, we show the results of experiments to see how much the operational stiffness of the arm changes with $f^{const}$.
  \figref{figure:variable-stiffness} shows the operational stiffness ellipsoid of the left arm in the sagittal plane when $\bm{\theta}^{ref}$ is set as the state with the elbow of the left arm of Musashi-W bent at 90 degrees and $f^{const}$ is changed to $\{10, 30, 50, 70\}$ [N].
  The graph shows the displacement of the hand when a force of 1 N is applied from all directions.
  It can be seen that the size of the stiffness ellipsoid changes greatly with the change in $f^{const}$.
  Therefore, we use $f^{const}$ as the body stiffness value $\bm{k}^{ref}$ in this study.
}%
{%
  本研究で扱う可変剛性制御可能な筋骨格ヒューマノイドについて, その制御入力である$\bm{\theta}^{ref}$, $\bm{k}^{ref}$の詳細を述べる.
  本研究では以下のような関節角度$\bm{\theta}$, 筋張力$\bm{f}$, 筋長$\bm{l}$の関係を用いる\citep{kawaharazuka2018bodyimage}.
  \begin{linenomath}
    \begin{align}
      \bm{l} = \bm{h}_{bodyimage}(\bm{\theta}, \bm{f}) \label{eq:bodyimage}
    \end{align}
  \end{linenomath}
  この写像$\bm{h}_{bodyimage}$は実機におけるデータを使って学習により獲得される.
  この学習されたネットワークを使って制御をする際は, 指令関節角度$\bm{\theta}^{ref}$, 指令筋張力$\bm{f}^{ref}$を決め, これに対応した筋長$\bm{l}^{ref}=\bm{h}_{bodyimage}(\bm{\theta}^{ref}, \bm{f}^{ref})$を計算する.
  ただし, この値は測定されるべき筋長であるため, 実際には, 筋剛性制御\citep{shirai2011stiffness}におけるソフトウェアの筋の伸びを考慮した$\bm{l}^{send}$が最終的に実機に送られる\citep{kawaharazuka2018bodyimage}.

  ここで, $\bm{f}^{ref}$の与え方について考える.
  通常$\bm{f}^{ref}$は実現したい$\bm{\theta}^{ref}$に必要なだけの値を入力すべきである.
  一方\citep{kawaharazuka2018bodyimage}では, 初めに全筋に対する$\bm{f}^{ref}$に一定の値$f^{const}$を与え, ある程度指令関節角度を実現した後, その際に発揮されている筋張力$\bm{f}$を$\bm{f}^{ref}$としてもう一度入力することで, より正確な指令関節角度を実現していた.
  そして, この最初に送る$f^{const}$の値によってある程度身体剛性を変化させることができる.
  また, 本研究では最終的に$\bm{s}$と$\bm{u}$の関係は学習によって獲得されるため, $\bm{\theta}^{ref}$は正確に実現する必要はない.
  よって, 本研究の$\bm{u}$に含まれる身体剛性値$\bm{k}^{ref}$として$f^{const}$を用いることとし, $\bm{l}^{ref}=\bm{h}_{bodyimage}(\bm{\theta}^{ref}, \bm{f}^{const})$によりロボットを動作させることとする.
  この$f^{const}$を変化させながら学習用データを取得することで, 身体剛性値を考慮した動的布操作が可能となる.

  なお, 以下に実際に$f^{const}$によってどの程度手先の作業空間剛性が変化するのかを実験した結果を載せる.
  Musashi-Wの左腕の肘を90度に曲げた状態の$\bm{\theta}^{ref}$において, $f^{const}$を$\{10, 30, 50, 70\}$ [N]に変化させた際の, 矢状面における手先の剛性楕円を\figref{figure:variable-stiffness}に示す.
  グラフは, 全方向から力を1 Nかけた時の手先変位を表している.
  $f^{const}$の変化によって大きく手先の剛性楕円の大きさが変化していることがわかる.
  よって本研究では剛性値$\bm{k}^{ref}$として$f^{const}$を用いる.
}%

\subsection{Training of DPMPB} \label{subsec:training}
\switchlanguage%
{%
  By setting $\bm{u}$ randomly, or operating the robot with GUI or VR device, we collect the data of $\bm{s}$ and $\bm{u}$.
  For a single, coherent manipulation trial $k$, performed with the same cloth, the data $D_k=\{(\bm{s}_1, \bm{u}_{1}), (\bm{s}_2, \bm{u}_2), \cdots, (\bm{s}_{T_{k}}, \bm{u}_{T_{k}})\}$ ($1 \leq k \leq K$, where $K$ is the total number of trials and $T_{k}$ is the number of time steps for the trial $k$).
  Then, we obtain the data $D_{train}=\{(D_1, \bm{p}_1), (D_2, \bm{p}_2), \cdots, (D_{K}, \bm{p}_K)\}$ for training.
  $\bm{p}_k$ is the parametric bias for the trial $k$, a variable that has a common value during that one trial and a different value for different trials.
  We use this data $D_{train}$ to train the DPMPB.
  In the usual training, only the network weight $W$ is updated, but in this study, $W$ and $\bm{p}_{k}$ are updated simultaneously as below,
  \begin{align}
    W \gets W - \beta\frac{\partial{L}}{\partial{W}} \label{eq:update-w}\\
    \bm{p}_{k} \gets \bm{p}_{k} - \beta\frac{\partial{L}}{\partial{\bm{p}_{k}}} \label{eq:update-p}
  \end{align}
  where $L$ is the loss function (mean squared error, in this study) and $\beta$ is the learning rate.
  $\bm{p}_{k}$ will be embedded with the difference in dynamics in each trial, i.e. the dynamics of the manipulated cloth in this study.
  $\bm{p}_{k}$ is trained with an initial value of $\bm{0}$, and it is not necessary to directly provide parameters related to the dynamics of the cloth.
  Therefore, parametric bias can be applied to cases where the dynamics parameters are not known, as in the handling of actual cloth.
}%
{%
  $\bm{u}$をランダムに送る, または人間がロボットをGUIやVRデバイスにより操作し, その際の$\bm{s}$と$\bm{u}$のデータを集めていく.
  同じ布を持った状態で行った, ある一回のまとまったマニピュレーションの試行$k$について, データ$D_k=\{(\bm{s}_1, \bm{u}_{1}), (\bm{s}_2, \bm{u}_2), \cdots, (\bm{s}_{T_{k}}, \bm{u}_{T_{k}})\}$を得る($1 \leq k \leq K$, $K$は全試行回数, $T_{k}$はその試行$k$に関する動作ステップ数とする).
  そして, 学習に用いるデータ$D_{train}=\{(D_1, \bm{p}_1), (D_2, \bm{p}_2), \cdots, (D_{K}, \bm{p}_K)\}$を得る.
  $\bm{p}_k$は試行$k$に関するParametric Biasであり, その一回の試行中については共通の値で, 異なる試行については別の値となる変数である.
  このデータ$D_{train}$を用いてDPMPBを学習させる.
  通常の学習ではネットワークの重み$W$が更新されるが, 本研究では$W$と$\bm{p}_{k}$を以下のように同時に更新していく.
  \begin{align}
    \bm{W} \gets \bm{W} - \beta\frac{\partial{L}}{\partial{W}} \label{eq:update-w}\\
    \bm{p}_{k} \gets \bm{p}_{k} - \beta\frac{\partial{L}}{\partial{\bm{p}_{k}}} \label{eq:update-p}
  \end{align}
  ここで, $L$は損失関数, $\beta$は学習率を表す.
  これにより, $\bm{p}_{k}$にはそれぞれの試行におけるダイナミクスの違い, つまり, 本研究では操作した布のダイナミクスが埋め込まれることになる.
  この$\bm{p}_{k}$は初期値を0として$W$と同時に学習され, 布のダイナミクスに関するパラメータを直接与える必要はない.
  そのため, 実際の布のようにダイナミクスのパラメータがわからないような場合にも適用することが可能である.
  なお, 学習の際の損失関数はmean squared errorとした.
}%

\subsection{Online Estimation of Cloth Material} \label{subsec:online-estimation}
\switchlanguage%
{%
  When manipulating a new cloth, if we do not know its correct dynamics, we will not be able to perform the manipulation correctly.
  Therefore, we need to obtain data on how the shape of the cloth changes when the cloth is manipulated, and estimate the dynamics of the cloth, i.e. $\bm{p}$ in this study, based on this data.
  First, we obtain the data $D_{new}$ ($D_{train}$ for one trial with $K=1$) as in \secref{subsec:training} when the cloth is manipulated by random commands, GUI, or the controller described later (\secref{subsec:control}).
  Using this data, we update only $\bm{p}$ while $W$ is fixed, unlike the training process of \secref{subsec:training}.
  We do not execute \equref{eq:update-w}, but update only $\bm{p}$ by \equref{eq:update-p} using the current $\bm{p}$ as the initial value.
  Since $\bm{p}$ is of very low dimension compared to $W$, the overfitting problem can be prevented.
  In other words, by updating only the terms related to the cloth dynamics, we make the dynamics of DPMPB consistent with $D_{new}$.
  In this case, the update rule is MomentumSGD.
}%
{%
  新しい布を操作するとき, そのダイナミクスがわからなければ, 正しくマニピュレーションを実行することはできない.
  そこで, 少し布を動かした際に, どのように布の形状が変化するのかのデータを取得し, これを元に, 布のダイナミクス, つまり$\bm{p}$を推定する必要がある.
  まず, \secref{subsec:training}と同様にその布をランダムやGUI, 後に述べる制御器(\secref{subsec:control})により操作したときのデータ$D_{new}$ ($D_{train}$において$K=1$とした一試行分のデータを表す)を得る.
  このデータを用いて, 学習時とは異なり, $W$を固定したうえで, $\bm{p}$のみ更新していく.
  具体的には, \equref{eq:update-w}は行わず, 現在の$\bm{p}$を初期値として, \equref{eq:update-p}により$\bm{p}$を更新していく.
  布のダイナミクスに関する項だけを更新することで, DPMPBのダイナミクスを$D_{new}$に合致させていく.
  $\bm{p}$は$W$に比べて非常に低次元であるため, 過学習を防ぐことができる.
  このときの更新則はMomentumSGDとした.
}%

\subsection{Dynamic Cloth Manipulation using DPMPB} \label{subsec:control}
\switchlanguage%
{%
  We describe the dynamic cloth manipulation control using DPMPB.
  First, we obtain the target image $I^{ref}$, which is the target state of the cloth, and compress it to $\bm{z}^{ref}$ by AutoEncoder.
  $\bm{f}^{ref}$ and $\bm{l}^{ref}$ are set to $\bm{0}$, and $\bm{s}^{ref}$ is generated by combining them.
  Next, we determine the number of expansions of DPMPB $N^{control}_{seq}$ and set the initial value $\bm{u}^{init}_{seq}$ as $\bm{u}^{opt}_{seq}$, which is $N^{control}_{seq}$ steps of $\bm{u}$ to be optimized (the abbreviation of $\bm{u}^{opt}_{[t, t+N^{control}_{seq}-1]}$).
  We calculate the loss function as follows, and update $\bm{u}^{opt}_{seq}$ from this value while $W$ and $\bm{p}$ are fixed.
  \begin{linenomath}
    \begin{align}
      L &= h_{loss}(\bm{s}^{ref}_{seq}, \bm{s}^{pred}_{seq}) \label{eq:loss}\\
      \bm{g} &= \frac{\partial{L}}{\partial{\bm{u}^{opt}_{seq}}} \label{eq:gradient}\\
      \bm{u}^{opt}_{seq} &\gets \bm{u}^{opt}_{seq} - \gamma\frac{\bm{g}}{||\bm{g}||_{2}} \label{eq:backprop}
    \end{align}
  \end{linenomath}
  where $h_{loss}$ is the loss function (to be explained later), $\bm{s}^{ref}_{seq}$ is a time series of the target cloth states with $\bm{s}^{ref}$ arranged in $N^{control}_{seq}$ steps, $\bm{s}^{pred}_{seq}$ is $\bm{s}^{pred}_{[t+1, t+N^{control}_{seq}]}$, which is the time series of $\bm{s}$ predicted when $\bm{u}^{opt}_{seq}$ is applied to the current state $\bm{s}_{t}$, $||\cdot||_{2}$ is L2 norm, and $\gamma$ is the learning rate.
  \equref{eq:gradient} can be computed by the backpropagation method of neural networks, and \equref{eq:backprop} represents the gradient descent method.
  In other words, the time series control command is updated to make the predicted time series of $\bm{s}$ closer to the target value.
  Note that $W$ is obtained in \secref{subsec:training}, and $\bm{p}$ is obtained in \secref{subsec:online-estimation}.
  Here, $\gamma$ can be constant, but in this study, we update $\bm{u}^{opt}_{seq}$ using multiple values of $\gamma$ for faster convergence.
  We use the value obtained by dividing $[0, \gamma_{max}]$ into $N^{control}_{batch}$ equal logarithmic intervals ($\gamma_{max}$ is the maximum value of $\gamma$) as $\gamma$, update $\bm{u}^{opt}_{seq}$ using each $\gamma$, and adopt $\bm{u}^{opt}_{seq}$ with the smallest $L$ among them, repeating the process $N^{control}_{iter}$ times.
  An appropriate $\gamma$ is selected for each iteration and \equref{eq:backprop} is performed.
  Also, for the initial value $\bm{u}^{init}_{seq}$ of $\bm{u}^{opt}_{seq}$, we use the value optimized in the previous step $\bm{u}^{prev}_{seq}$ (the abbreviation of $\bm{u}^{prev}_{[t, t+N^{control}_{seq}-1]}$) by shifting it one step to the left and duplicating the last term, $\bm{u}^{prev}_{\{t+1, \cdots, t+N^{control}_{seq}-1, t+N^{control}_{seq}-1\}}$.
  This allows us to achieve faster convergence, taking into account the previous optimization results.
  The final obtained target value $\bm{u}^{opt}_{t}$ at the current time step of $\bm{u}^{opt}_{seq}$ is sent to the actual robot.

  Here, we consider dynamic cloth manipulation tasks such as spreading out a bed sheet or a picnic sheet in the air.
  Let $\bm{s}^{ref}$ be the target value that contains the image $\bm{z}^{ref}$ of the unfolded state of the cloth, and we consider trying to achieve the target value.
  In this case, if the cloth does not spread well at the first attempt, it would be not possible to make minor adjustments, and it is necessary to generate a series of motions to try to spread the cloth from the beginning.
  In other words, even if the loss is lowered to some extent by the first attempt, it is necessary to go through a state where the loss is increased by large motions in order to lower the loss further.
  Therefore, if we just define $\bm{s}^{ref}_{seq}$ as a vector of the same $\bm{s}^{ref}$, we end up with $\bm{u}^{opt}_{seq}$ that hardly moves after the first attempt.
  The same problem occurs when $\bm{s}^{ref}$ is defined as the state of spreading a cloth in the air, since that state is not always realized.
  Thus, it is necessary to realize the target state by periodically spreading the cloth over and over again.
  Based on this feature, in this study, we determine the number of periodic steps of the periodic motion, $N^{control}_{periodic}$, and change $h_{loss}$ for each period.
  In this study, $h_{loss}$ is expressed as follows.
  \begin{linenomath}
    \begin{align}
      h_{loss}(\bm{s}^{ref}_{seq}, \bm{s}^{pred}_{seq}) = ||\bm{m}_{t}\otimes(\bm{z}^{ref}_{seq}-\bm{z}^{pred}_{seq})||_{2} + w_{loss}||\bm{f}^{pred}_{seq}||_{2} \label{eq:loss-detailed}
    \end{align}
  \end{linenomath}
  where $\{\bm{z}, \bm{f}\}^{\{ref, pred\}}_{seq}$ is the value of $\{\bm{z}, \bm{f}\}$ extracted from $\bm{s}^{\{ref, pred\}}_{seq}$, and $w_{loss}$ is the weight constant.
  $\bm{m}_{t}$ ($\in \{0, 1\}^{N^{control}_{seq}}$) is a vector with 1 occurring at every $N^{control}_{periodic}$ step and 0 otherwise.
  It is shifted to the left at each time step, and 0 or 1 is inserted from the right depending on $N^{control}_{periodic}$ (e.g. if $N^{control}_{seq}=N^{control}_{periodic}=4$, then $(0, 1, 0, 0 ) \rightarrow (1, 0, 0, 0) \rightarrow (0, 0, 0, 1)$ in this order).
  This makes it possible to bring the cloth state $\bm{z}$ closer to the target value every $N^{control}_{periodic}$ step, and enables dynamic cloth manipulation that handles periodic states which can only be realized momentarily.
  The second term on the right-hand side of \equref{eq:loss-detailed} is a term to minimize muscle tension as much as possible.

  In this study, we set $N^{control}_{seq}=8$, $N^{control}_{batch}=30$, $\gamma_{max}=1.0$, $N^{control}_{iter}=3$, $w_{loss}=0.001$, and $N^{control}_{periodic}=8$.
}%
{%
  DPMPBを使った動的布操作制御について述べる.
  まず, 布の指令状態となる画像$I^{ref}$を取得し, AutoEncoderにより$\bm{z}^{ref}$に圧縮する.
  $\bm{f}^{ref}$と$\bm{l}^{ref}$は$\bm{0}$とし, これらを合わせて$\bm{s}^{ref}$を生成する.
  次に, DPMPBの展開数$N^{control}_{seq}$を決め, 最適化すべき$\bm{u}$の$N^{control}_{seq}$ステップの時系列$\bm{u}^{opt}_{seq}$ ($\bm{u}^{opt}_{[t, t+N^{control}_{seq}-1]}$の略)に初期値$\bm{u}^{init}_{seq}$を設定する.
  以下のように, 損失関数を計算し$\bm{u}^{opt}_{seq}$を更新していく.
  \begin{linenomath}
    \begin{align}
      L &= h_{loss}(\bm{s}^{ref}_{seq}, \bm{s}^{pred}_{seq}) \label{eq:loss}\\
      \bm{g} &= \frac{\partial{L}}{\partial{\bm{u}^{opt}_{seq}}}\\
      \bm{u}^{opt}_{seq} &\gets \bm{u}^{opt}_{seq} - \gamma\frac{\bm{g}}{||\bm{g}||_{2}} \label{eq:backprop}
    \end{align}
  \end{linenomath}
  ここで, $h_{loss}$は損失関数(後に説明する), $\bm{s}^{ref}_{seq}$は$\bm{s}^{ref}$を$N^{control}_{seq}$ステップ並べた指令布状態の時系列, $\bm{s}^{pred}_{seq}$は現在の状態$\bm{s}_{t}$に$\bm{u}^{opt}_{seq}$を順に入れていったときに予測された$\bm{s}$の時系列$\bm{s}^{pred}_{[t+1, t+N^{control}_{seq}]}$, $||\cdot||_{2}$はL2ノルム, $\gamma$は学習率を表す.
  \equref{eq:gradient}はニューラルネットワークの誤差逆伝播法により計算可能であり, \equref{eq:backprop}は勾配降下法を表している.
  つまり, 予測される$\bm{s}$の時系列を指令値に近づけるように時系列の制御入力を更新していく.
  なお, $W$は\secref{subsec:training}で得られたもの, $\bm{p}$は\secref{subsec:online-estimation}で得られたものを使う.
  ここで, $\gamma$は一定値でも良いが, 本研究ではより早い収束性のため, 複数の$\gamma$を使って$\bm{u}^{opt}_{seq}$を更新する.
  $\gamma$を$[0, \gamma_{max}]$を対数的に等間隔に$N^{control}_{batch}$等分した値とし($\gamma_{max}$は$\gamma$の最大値), それぞれの$\gamma$を使って$\bm{u}^{opt}_{seq}$を更新し, その中で最も$L$の小さかった$\bm{u}^{opt}_{seq}$を採用するという処理を$N^{control}_{iter}$回繰り返す.
  つまり, 毎イテレーションごとに適切な$\gamma$が選択され, \equref{eq:backprop}が行われることになる.
  また, $\bm{u}^{opt}_{seq}$の初期値$\bm{u}^{init}_{seq}$は, 前ステップで最適化された値$\bm{u}^{prev}_{seq}$ ($\bm{u}^{prev}_{[t, t+N^{control}_{seq}-1]}$の略)を一ステップ分左へシフトし, 最後の項を複製したもの$\bm{u}^{prev}_{\{t+1, \cdots, t+N^{control}_{seq}-1, t+N^{control}_{seq}-1\}}$を用いる.
  これにより, これまでの最適化結果を考慮したより早い収束が得られる.
  最終的に得られた$\bm{u}^{opt}_{seq}$の現在のタイムステップにおける指令値$\bm{u}^{opt}_{t}$を実機に送る.

  ここで, 具体的にベッドメイキングの際にシーツを大きく広げたりレジャーシートを空中で広げたいする動的布操作について考える.
  この布を広げた状態の画像$\bm{z}^{ref}$を含む指令値を$\bm{s}^{ref}$として, これを実現しようとすることを考える.
  このとき, もし一度で布を広げることがうまくいかなかった場合, そこから微修正ができるわけではなく, また最初から布を手前にもってきて一気に開くような一連の動作を生成する必要がある.
  つまり, 一度最初の動きで損失がある程度下がっても, さらに損失を下げるためにはもう一度大きく身体を動かし損失を大きくする状態を経由して動作をやり直す必要がある.
  そのため, $\bm{s}^{ref}$を並べたベクトルを$\bm{s}^{ref}_{seq}$としただけでは, 一度大きく広げようとした後は, ほとんど動かない$\bm{u}^{opt}_{seq}$が生成されてしまう.
  また, 空中で布を広げるような状態を$\bm{s}^{ref}$とした場合は, その状態が常に実現できるわけではないため, 同様の問題が起きる.
  つまり, 何度も周期的に広げる動作を行うことで, 指令状態を実現する必要がある.
  この特徴を踏まえたうえで, 本研究では周期的運動の周期ステップ数$N^{control}_{periodic}$を決め, その周期ごとに$h_{loss}$を変化させる.
  本研究では以下のように$h_{loss}$を表現する.
  \begin{linenomath}
    \begin{align}
      h_{loss}(\bm{s}^{ref}_{seq}, \bm{s}^{pred}_{seq}) = ||\bm{m}_{t}\otimes(\bm{z}^{ref}_{seq}-\bm{z}^{pred}_{seq})||_{2} + w_{loss}||\bm{f}^{pred}_{seq}||_{2} \label{eq:loss-detailed}
    \end{align}
  \end{linenomath}
  ここで, $\{\bm{z}, \bm{f}\}^{\{ref, pred\}}_{seq}$は$\bm{s}^{\{ref, pred\}}_{seq}$の$\{\bm{z}, \bm{f}\}$のみを抜き出したもの, $w_{loss}$は重み付けの係数である.
  $\bm{m}_{t}$ ($\in \{0, 1\}^{N^{control}_{seq}}$)は$N^{control}_{periodic}$ステップごとに1が出現し, その他は0のベクトルである.
  $\bm{m}_{t}$は一ステップごとに左へシフトし, 右からは$N^{control}_{periodic}$に応じて0または1が挿入される(e.g. もし$N^{control}_{seq}=N^{control}_{periodic}=4$ならば順に$(0, 1, 0, 0) \rightarrow (1, 0, 0, 0) \rightarrow (0, 0, 0, 1)$になる).
  これにより, $N^{control}_{periodic}$ごとに布の状態$\bm{z}$を指令値に近づけることができ, 周期的かつ, 一瞬しか実現できない状態を扱う動的布操作が可能となる.
  \equref{eq:loss-detailed}の右辺第二項は, なるべく大きな筋張力は出さないようにするための項である.

  本研究では, $N^{control}_{seq}=8$, $N^{control}_{batch}=30$, $\gamma_{max}=1.0$, $N^{control}_{iter}=3$, $w_{loss}=0.001$, $N^{control}_{periodic}=8$とした.
}%

\begin{figure}[t]
  \centering
  \includegraphics[width=0.8\columnwidth]{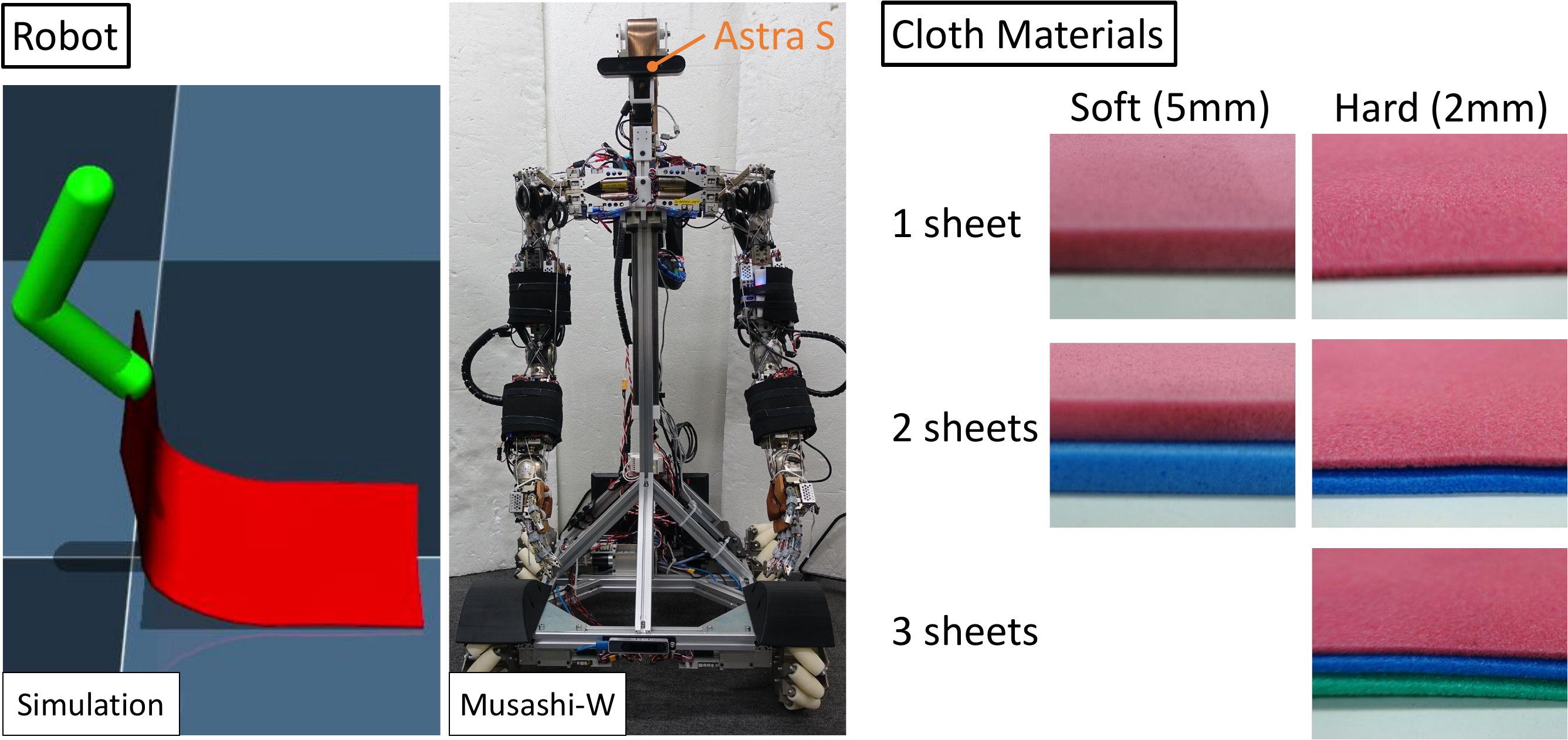}
  \caption{Experimental setup: the simulated simple robot and the musculoskeletal humanoid Musashi-W used in this study and cloth materials of soft and hard type polyethylene foam.}
  \label{figure:exp-setup}
\end{figure}

\section{Experiments} \label{sec:experiments}
\subsection{Experimental Setup} \label{subsec:exp-setup}
\switchlanguage%
{%
  The robot and the types of cloth used in this study are shown in \figref{figure:exp-setup}.
  First, we conduct a cloth manipulation experiment using a simple 2-DOF robot in simulation with Mujoco \citep{todorov2012mujoco}.
  Here, the cloth is modeled as a collection of 3$\times$6 point masses, and the damping $C_{damp}$ between those point masses and the weight $C_{mass}$ of the entire cloth can be changed.
  Next, we use the actual robot Musashi-W, which is a musculoskeletal dual arm robot MusashiDarm \citep{kawaharazuka2019musashi} with a mechanum wheeled base and a z-axis slider.
  The head is equipped with Astra S camera (Orbbec 3D Technology International, Inc.).
  In this study, we use two kinds of cloth: soft type (5 mm thick) and hard type (2 mm thick) of polyethylene foam (TAKASHIMA color sheet, WAKI Factory, Inc.).
  Polyethylene foam was used instead of actual cloth in this study due to the joint speed limitation of Musashi-W.
  For the soft type, we prepare one or two sheets, and for the hard type, we prepare one, two, or three sheets (denoted as soft-1, soft-2, hard-1, hard-2, hard-3).
  % A metal rod is tied to one side of the cloth to make it easier to hold.

  We move the arms of the simulated robot and Musashi-W only in the sagittal plane, with two degrees of freedom in the shoulder and elbow pitch axes.
  Although the the control command of the simulated robot is two dimensional without hardware body stiffness, the control command of Musashi-W is three dimensional including the stiffness (hardware stiffness was difficult to reproduce in simulation).
  The image is compressed by AutoEncoder after color extraction, binarization, closing opening, and resizing (the front part of the cloth is red and the back part is a different color such as blue or black).
}%
{%
  本研究で用いるロボット・布の種類を\figref{figure:exp-setup}に示す.
  本研究ではまず検証として, Mujoco \citep{todorov2012mujoco}を用いたシミュレーションにおいて, 簡単な2自由度ロボットを使った布操作実験を行う.
  この際, 布は3$\times$6の質点の集まりとしてモデル化されており, それら質点の間のdamping $C_{damp}$と布全体の重さ$C_{mass}$を変更することができる.
  次に, 筋骨格双腕MusashiDarm \citep{kawaharazuka2019musashi}にメカナム台車とz軸のスライダーが追加されたMusashi-Wを用いて実験を行う.
  頭部にはカメラとして, Astra S (Orbbec 3D Technology International, Inc.)がついている.
  本研究ではポリエチレンフォームのsoft type (厚さ5 mm)とhard type (厚さ2 mm) (TAKASHIMA color sheet, WAKI Factory, Inc.)の二種類の素材を布として扱う.
  本来であれば布を用いるべきであるが, Musashi-Wの関節速度制限から今回はポリエチレンフォームを用いた.
  なお, soft typeは1枚または2枚, hard typeは1枚, 2枚, または3枚重ねたものを用意する(soft-1, soft-2, hard-1, hard-2, hard-3のように表記する).

  SimulationとMusashi-Wの腕は矢状面でのみの動作とし, 肩と肘のpitch軸2自由度のみを動かす.
  Simulationについてはハードウェアの剛性値に関するパラメータはないが, Musashi-Wについては剛性値を含めて制御入力は3自由度である(simulationにおいてハードウェア剛性を再現することは難しかった).
  画像は布の赤(表部分は赤, 裏は青や黒の異なる色になっている)を色抽出・2値化・拡大・縮小・リサイズを施した後, AutoEncoderにより圧縮される.
}%

\begin{figure}[t]
  \centering
  \includegraphics[width=0.6\columnwidth]{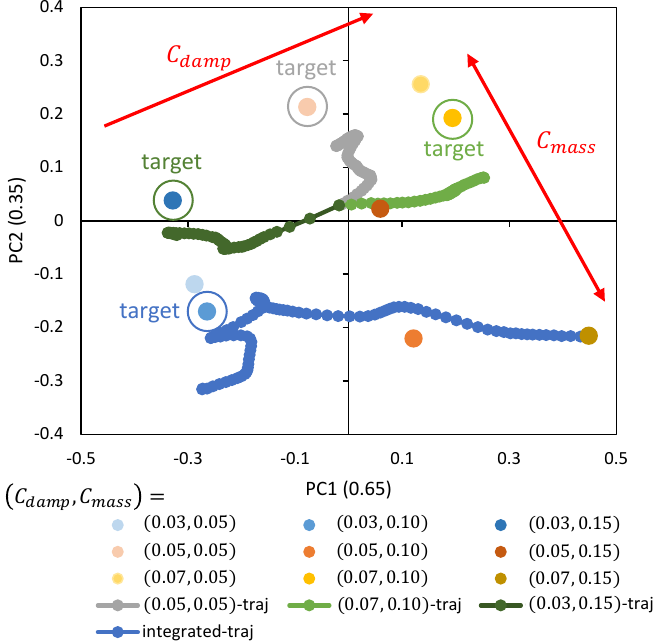}
  \caption{Simulation experiment: the trained parametric bias when setting $C_{damp}=\{0.03, 0.05, 0.07\}$ and $C_{mass}=\{0.05, 0.10, 0.15\}$, the trajectory of online updated parametric bias when setting $(C_{damp}, C_{mass})=\{(0.05, 0.05), (0.07, 0.10), (0.03, 0.15)\}$, and the trajectory of online updated parametric bias when setting $(C_{damp}, C_{mass})=(0.03, 0.10)$ for the integrated experiment.}
  \label{figure:sim-pb}
\end{figure}

% \begin{figure}[t]
%   \centering
%   \includegraphics[width=0.8\columnwidth]{figs/sim-control}
%   \caption{Simulation experiment: the rate (y-axis) of $||\bm{z}^{ref}-\bm{z}||_{2} <$threshold (x-axis) when conducting experiments of dynamic manipulation of two cloths Target-1 and Target-2 when setting $(C_{damp}, C_{mass})=\{(0.03, 0.05), (0.07, 0.15)\}$ regarding Control and Random.}
%   \label{figure:sim-control}
% \end{figure}

\begin{figure}[t]
  \centering
  \includegraphics[width=0.8\columnwidth]{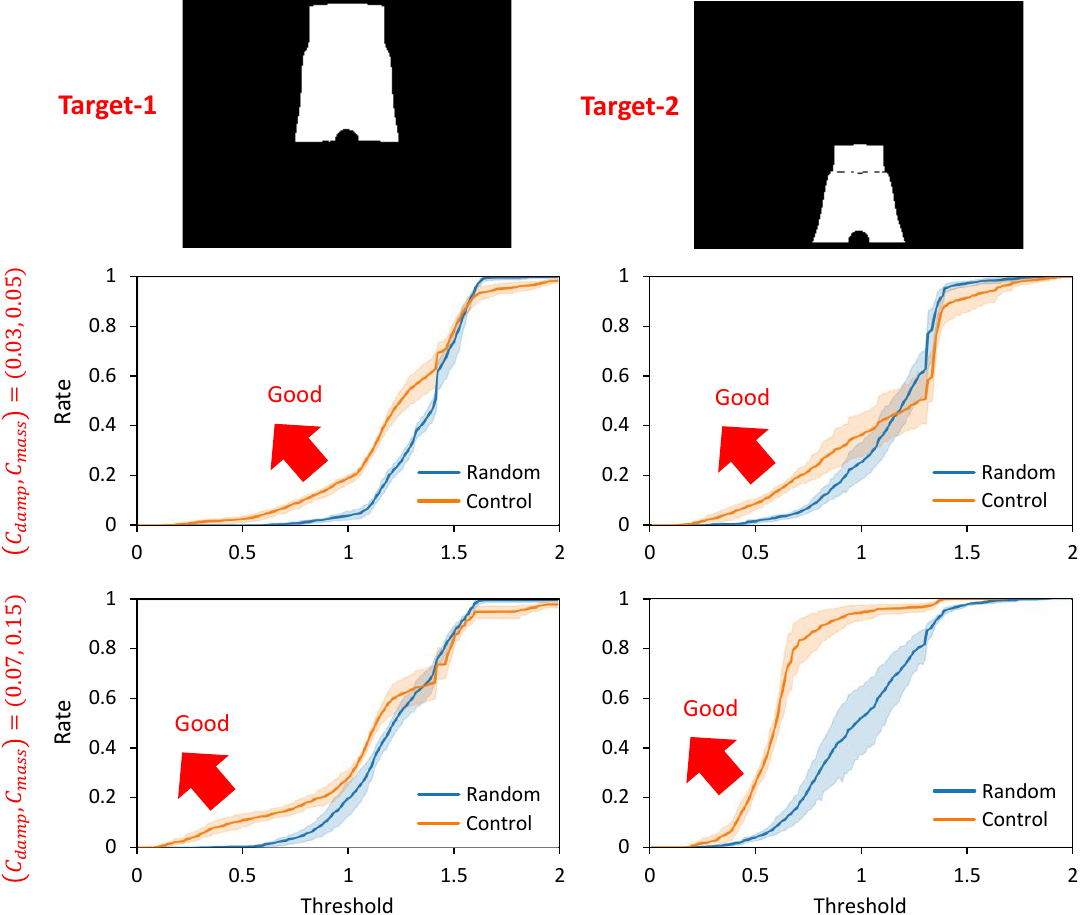}
  \caption{Simulation experiment: the rate (y-axis) of $||\bm{z}^{ref}-\bm{z}||_{2} <$threshold (x-axis) when conducting experiments of dynamic manipulation of two cloths Target-1 and Target-2 when setting $(C_{damp}, C_{mass})=\{(0.03, 0.05), (0.07, 0.15)\}$ regarding Control and Random.}
  \label{figure:sim-control}
\end{figure}

\begin{figure}[t]
  \centering
  \includegraphics[width=0.6\columnwidth]{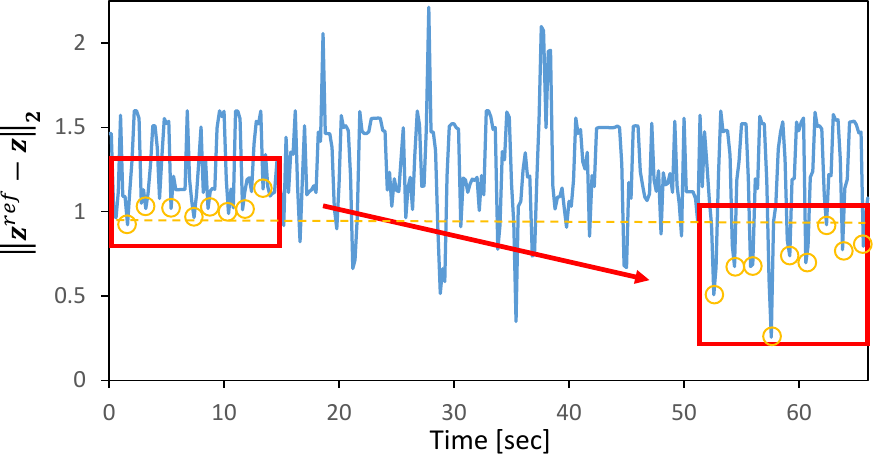}
  \caption{Simulation experiment: the transition of $||\bm{z}^{ref}-\bm{z}||_{2}$ for the integrated experiment of online estimation and dynamic cloth manipulation.}
  \label{figure:sim-integrated}
\end{figure}

\begin{figure}[t]
  \centering
  \includegraphics[width=0.4\columnwidth]{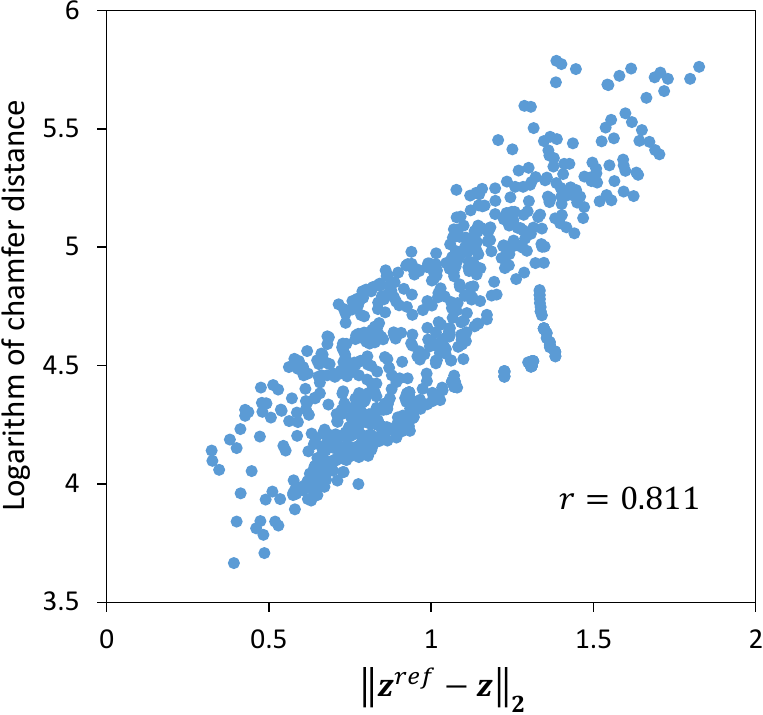}
  \caption{The correlation between logarithm of chamfer distance and $||\bm{z}^{ref}-\bm{z}||_{2}$.}
  \label{figure:sim-zchamfer}
\end{figure}

\subsection{Simulation Experiment} \label{subsec:exp-sim}
\switchlanguage%
{%
  The cloth characteristics are changed to $C_{damp}=\{0.03, 0.05, 0.07\}$ and $C_{mass}=\{0.05, 0.10, 0.15\}$ while randomly commanding $\bm{\theta}^{ref}$ (Random) for about 50 seconds, and while human operation by GUI (joint angle is given by GUI) for about 50 seconds.
  Using a total of 4500 data points, we trained DPMPB with the number of LSTM expansions set to 30, the number of batches being 300, and the number of epochs being 300 (these parameters are empirically set).
  Principle Component Analysis (PCA) is applied to the parametric bias for each cloth obtained in the training, and its arrangement in the two-dimensional plane is shown in \figref{figure:sim-pb}.
  Note that even if $\bm{p}$ is two-dimensional, the axes of the principal components, PC1 and PC2, become more defined by applying PCA.
  It can be seen that the PBs are neatly arranged mainly on the axis of PC1 according to the magnitude of $C_{damp}$.
  On the other hand, for the axis of PC2, some PBs are not arranged according to the magnitude of $C_{mass}$, and we can see that the dynamics change due to $C_{mass}$ is complex.
  Also, the contribution ratio in PCA is 0.65 for PC1 and 0.35 for PC2, indicating that the influence of $C_{mass}$ on the dynamics of the cloth is small.

  Here, with each cloth of $(C_{damp}, C_{mass})=\{(0.05, 0.05), (0.07, 0.10), (0.03, 0.15)\}$, the Random motion and the online material estimation in \secref{subsec:online-estimation} are executed for about 40 seconds.
  The trajectory of the parametric bias (traj) is shown in \figref{figure:sim-pb}.
  Note that the initial value of PB is $\bm{0}$ (the origin of PB is not necessarily at the origin on the figure, since PCA is applied).
  It can be seen that the current PB gradually approaches the respective PB values obtained during training in the dynamics of the current cloth.
  In particular, the accuracy of the online estimation is high for the axis of $C_{damp}$.
  On the other hand, for the axis of $C_{mass}$, though the material can be correctly recognized to some extent, the accuracy is less than for the axis of $C_{damp}$.

  Next, we performed the control in \secref{subsec:control} by setting the cloth characteristics to $(C_{damp}, C_{mass})=\{(0.03, 0.05), (0.07, 0.15)\}$ and by setting the current PB value to the PB obtained during the training of the object to be manipulated.
  Two images, Target-1 and Target-2 of \figref{figure:sim-control}, are given as the target images of the cloth.
  We denote the case of control in this study as Control and the random trial as Random.
  For each of the two types of cloth, the Control and Random trials are performed for 50 seconds each, using the two target images.
  The percentage (y-axis) of $||\bm{z}^{ref}-\bm{z}||_{2}$ lower than a certain threshold (x-axis) is shown in \figref{figure:sim-control}.
  \figref{figure:sim-control} indicates that the more the graph expands in the upper-left direction (the larger the y-axis value becomes when the x-axis value is low), the more the target image is realized.
  Here, in order to verify the reproducibility, all experiments including learning and control are conducted five times, and the mean and variance of these trials are shown in the graphs.
  Note that the mean of $||\bm{z}^{ref}-\bm{z}||_{2}$ in the initial condition shown in the left figure of \figref{figure:exp-setup} is 1.55 for Target-1 and 1.36 for Target-2.
  In all cases, Control outperforms Random and succeeds in realizing the target image accurately.
  Since Target-1 is more difficult than Target-2, especially when $(C_{damp}, C_{mass})=(0.03, 0.05)$, the difference is clearly shown.
  Since the larger $C_{damp}$ causes slower deformation of the cloth and makes it easier to lift up, the target image is relatively correctly realized when is set $(C_{damp}, C_{mass})$ to $(0.07, 0.15)$, compared to when it is set to $(0.03, 0.05)$.
  The reproducibility of performance with respect to learning and control is also high, especially in areas with a low threshold (x-axis), where the variance is small.

  Next, assuming that the current cloth properties are $(C_{damp}, C_{mass})=(0.03, 0.10)$ and the current PB values are $(C_{damp}, C_{mass})=(0.07, 0.15)$ obtained during training, we simultaneously perform the online material estimation in \secref{subsec:online-estimation} and control in \secref{subsec:control}.
  Note that the target image is set as Target-1 of \figref{figure:sim-control}.
  The transition of PB (integrated-traj) is shown in \figref{figure:sim-pb}, and the transition of $||\bm{z}^{ref}-\bm{z}||_{2}$ is shown in \figref{figure:sim-integrated}.
  It can be seen that the current PB gradually approaches the value obtained during training with $(C_{damp}, C_{mass})=(0.03, 0.10)$.
  Also, as PB becomes more accurate, $||\bm{z}^{ref}-\bm{z}||_{2}$ periodically shows smaller values.
  In other words, the online material estimation and the learning control can be performed together, and the control becomes more accurate as PB becomes more accurate.

  Finally, we show that $||\bm{z}^{ref}-\bm{z}||_{2}$ is close to the actual distance between the raw images, since the control and its evaluation are basically performed with the error of the latent variable $\bm{z}$ in this study.
  We use Symmetric Chamfer Distance \citep{borgefors1988chamfer}, which represents the similarity between binary images, to calculate the distance between the current raw image $I$ and its target value $I^{ref}$.
  The relationship between the logarithm of Chamfer Distance and $||\bm{z}^{ref}-\bm{z}||_{2}$ when the target image is set to Target-2 and the cloth is randomly moved for 90 seconds is shown in \figref{figure:sim-zchamfer}.
  The two values are well correlated, with a correlation coefficient of 0.811.
}%
{%
  布の特性を$C_{damp}=\{0.03, 0.05, 0.07\}$, $C_{mass}=\{0.05, 0.10, 0.15\}$と変更しながら$\bm{\theta}^{ref}$をランダムに指令した場合(Random)を約50秒間, GUI (関節角度をGUIで動かす)で人間が動作させた場合を約50秒間行いデータを取得した.
  計4500のデータを使い, LSTMの展開数を30, バッチ数を300, エポック数を300としてDPMPBを学習した (これらの値は実験的に決定している, empirically).
  その際に得られたそれぞれの布に関するParametric BiasにPrinciple Component Analysis (PCA)をかけ, 2次元平面内に配置した様子を\figref{figure:sim-pb}に示す.
  なお, たとえ$\bm{p}$が2次元でも, PCAを適用することで, $\bm{p}$の主成分PC1, PC2の軸がより鮮明になる.
  主にPC1の軸において, $C_{damp}$の大きさに従って綺麗にPBが配置されていることがわかる.
  一方で, PC2の軸について, 一部のPBは$C_{mass}$の大きさに沿って配置されているわけではなく, $C_{mass}$によるダイナミクスの変化は複雑であることが読み取れる.
  また, PCAにおける寄与率もPC1が0.65なのに対してPC2は0.35であり, 布のダイナミクスに対する$C_{mass}$の影響は小さいことがわかる.

  ここで, $(C_{damp}, C_{mass})=\{(0.05, 0.05), (0.07, 0.10), (0.03, 0.15)\}$のそれぞれの布を持った状態で, Randomの動作と\secref{subsec:online-estimation}のオンライン素材推定を約40秒間走らせる.
  このときのParametric Biasの軌跡(traj)を\figref{figure:sim-pb}に示す.
  なお, PBの初期値は$\bm{0}$とする(PCAをかけているため, PBの原点がグラフ上で原点にあるとは限らない).
  現在のPBは, 徐々に現在の布のダイナミクスにおいて訓練時に得られたそれぞれのPBの値へ近づいていくことがわかる.
  特に$C_{damp}$の軸についてはその精度は正確である.
  一方で, $C_{mass}$の軸についてはある程度正しく素材のパラメータを認識できているものの, その精度は$C_{damp}$の軸に比べると低いことがわかる.

  次に, 布の特性を$(C_{damp}, C_{mass})=\{(0.03, 0.05), (0.07, 0.15)\}$として, 現在PBの値を, 操作する物体の訓練時に得られたPBに設定した状態で, \secref{subsec:control}の制御を行う.
  このとき, 布の指令画像として\figref{figure:sim-control}の2つの画像Target-1とTarget-2を与える.
  また, 本研究の制御を行った場合をControl, ランダムな試行をRandomと表す.
  2種類の布について, 2種類の画像を指令値としてControlとRandomのそれぞれを50秒間行う.
  その際に, $||\bm{z}^{ref}-\bm{z}||_{2}$がある閾値(x軸)よりも低い割合(y軸)を\figref{figure:sim-control}に示す.
  \figref{figure:sim-control}はグラフが左上方向に膨らむほど, より正確に指令画像状態が実現できていることを表す.
  なお, 本実験については再現性検証のため, 全ての実験について学習から制御までを5回行い, その際の平均と分散をグラフに表示している.
  また, \figref{figure:exp-setup}の左図の初期状態における$||\bm{z}^{ref}-\bm{z}||_{2}$の平均は, Target-1について1.55, Target-2について1.36であった.
  どのケースについても, ControlはRandomを上回り, 指令画像を実現することに成功している.
  Target-1の方がTarget-2より難しいため, 特に$(C_{damp}, C_{mass})=(0.03, 0.05)$のときはその違いが如実に現れている.
  $C_{damp}$が大きい方が布の変形が遅くなり布が浮き上がりやすくなるため, $(C_{damp}, C_{mass})$が$(0.07, 0.15)$のときの方が, $(0.03, 0.05)$のときに比べて比較的指令画像を正しく実現できている.
  また, 学習・制御に関する性能の再現性も高く, 特にx軸の閾値が低い部分では分散が小さい.

  次に, 現在の布の特性を$(C_{damp}, C_{mass})=(0.03, 0.10)$, 現在のPB値を$(C_{damp}, C_{mass})=(0.07, 0.15)$で訓練時に得られた値として, \secref{subsec:online-estimation}の素材推定と\secref{subsec:control}の制御を同時に実行する実験を行う.
  なお, 指令画像は\figref{figure:sim-control}のTarget-1としている.
  この際のPBの遷移(integrated-traj)を\figref{figure:sim-pb}に, $||\bm{z}^{ref}-\bm{z}||_{2}$の遷移を\figref{figure:sim-integrated}に示す.
  現在のPBは徐々に$(C_{damp}, C_{mass})=(0.03, 0.10)$で訓練時に得られた値へと近づいていくことがわかる.
  また, PBが正確になるにつれて, $||\bm{z}^{ref}-\bm{z}||_{2}$が周期的により小さな値を示すようになる.
  つまり, オンラインの素材推定と本学習制御は両立して実行することが可能であり, PBが正確になるにつれて制御も正確になる.

  最後に, 本研究では基本的に潜在変数$\bm{z}$の誤差で制御とその評価を行っているが, この$||\bm{z}^{ref}-\bm{z}||_{2}$が実際の生画像同士の距離に近いことを示す.
  画像間の類似度を表すSymmetric Chamfer Distance \citep{borgefors1988chamfer}により, 2値画像における現在の生画像$I$とその指令値$I^{ref}$の距離を計算する.
  指令画像をTarget-2に設定し, 90秒間ランダムに布を動かした際のChamfer Distanceの対数と$||\bm{z}^{ref}-\bm{z}||_{2}$の関係を\figref{figure:sim-zchamfer}に示す.
  この2つの値は良く相関を示し, その相関係数は0.811となっている.
}%

\begin{figure}[t]
  \centering
  \includegraphics[width=0.6\columnwidth]{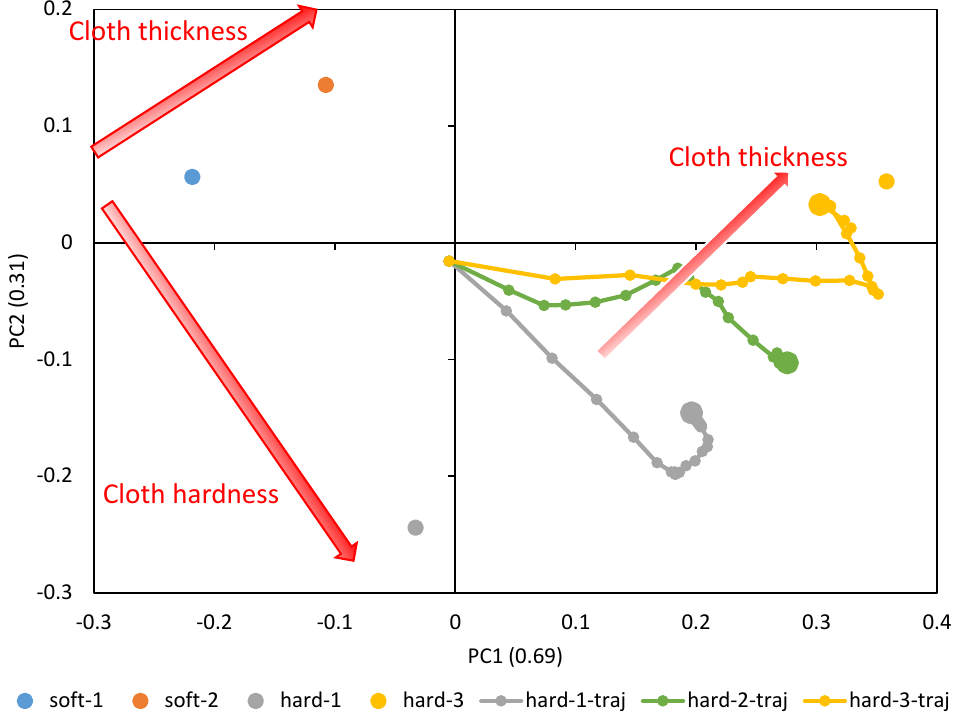}
  \caption{Actual robot experiment: the trained parametric bias for soft-1, soft-2, hard-1, and hard-3, and the trajectory of online updated parametric bias for hard-1, hard-2, and hard-3.}
  \label{figure:act-pb}
\end{figure}

\begin{figure}[t]
  \centering
  \includegraphics[width=0.90\columnwidth]{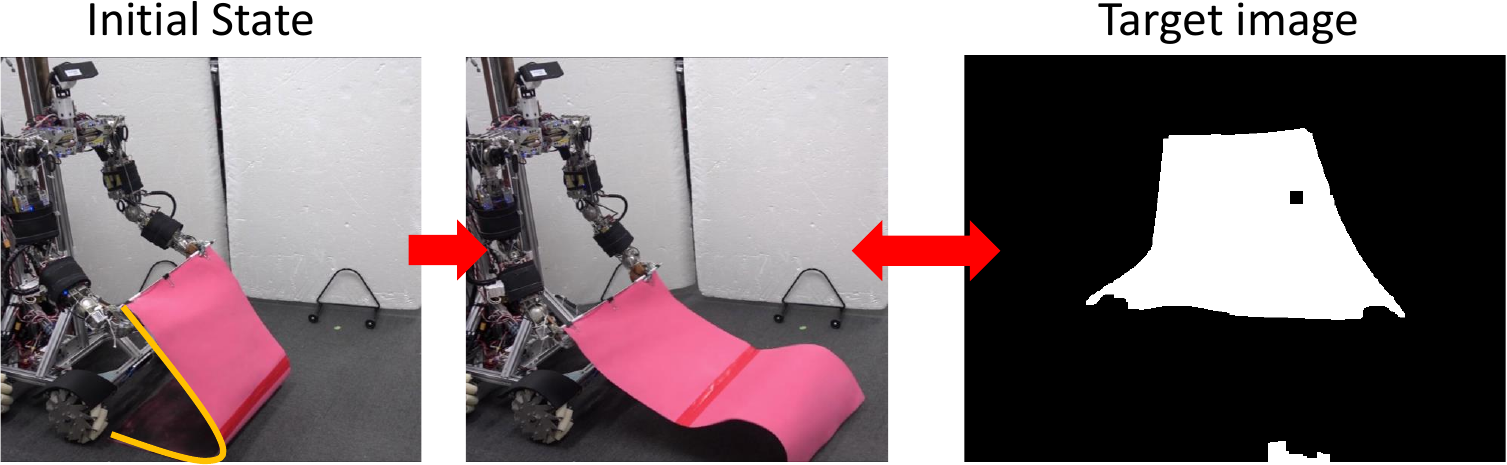}
  \caption{Experimental setup for the actual robot experiment: the initial state of cloth and target image of spread-out cloth.}
  \label{figure:exp-setup2}
\end{figure}

\begin{figure}[t]
  \centering
  \includegraphics[width=0.75\columnwidth]{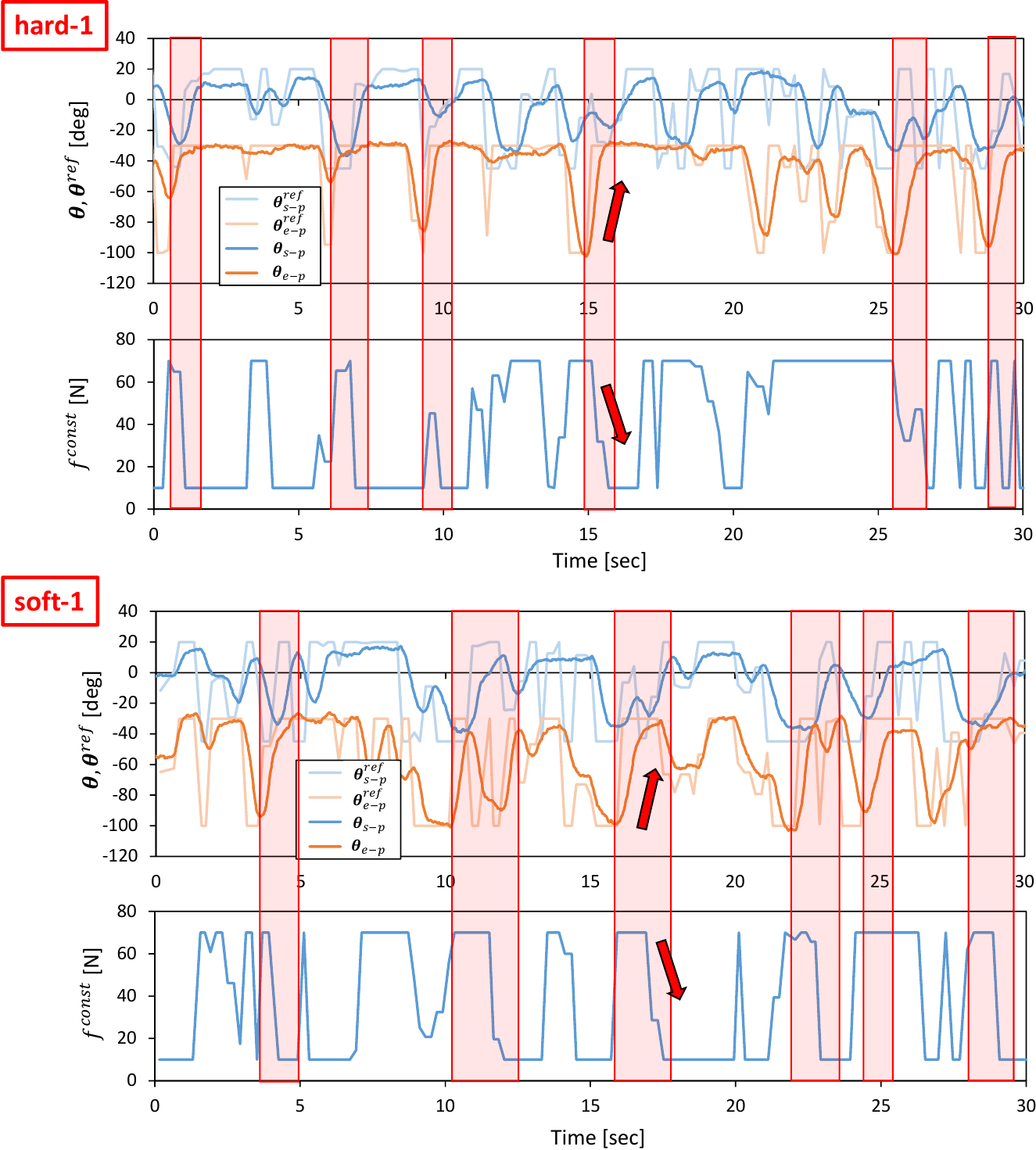}
  \caption{Actual robot experiment: examples of transitions of $\bm{\theta}^{ref}$, $\bm{\theta}$, $f^{const}$, and $||\bm{z}^{ref}-\bm{z}||_{2}$ when conducting an experiment of dynamic manipulation of hard-1 (upper graphs) and soft-1 (lower graphs).}
  \label{figure:test-control}
\end{figure}

\begin{figure}[t]
  \centering
  \includegraphics[width=0.5\columnwidth]{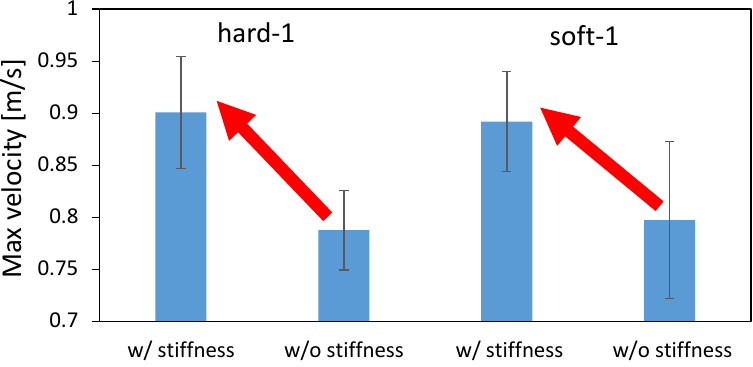}
  \caption{Actual robot experiment: the average and standard deviation of the maximum velocity of end effector when conducting experiments of dynamic manipulation of hard-1 and soft-1.}
  \label{figure:eval-vel}
\end{figure}

\begin{figure}[t]
  \centering
  \includegraphics[width=0.8\columnwidth]{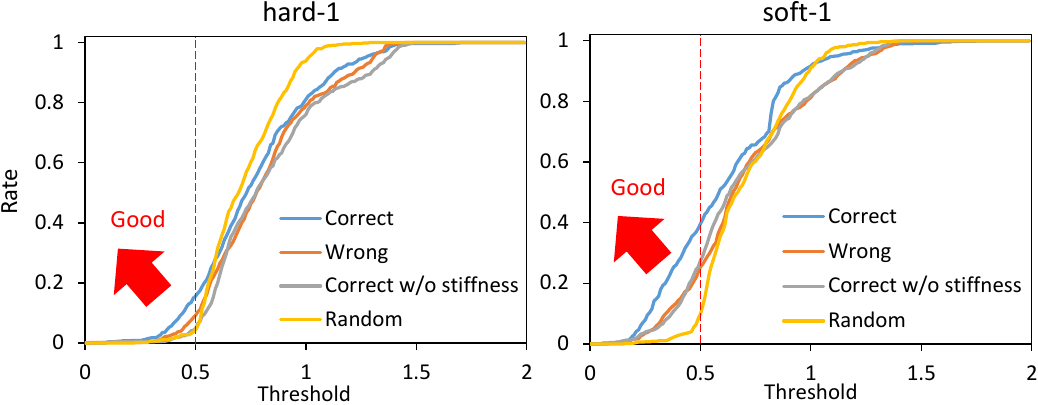}
  \caption{Actual robot experiment: the rate (y-axis) of $||\bm{z}^{ref}-\bm{z}||_{2} <$threshold (x-axis) when conducting experiments of dynamic manipulation of hard-1 and soft-1 regarding Correct, Wrong, Correct without stiffness, and Random.}
  \label{figure:eval-graph}
\end{figure}

\begin{figure}[t]
  \centering
  \includegraphics[width=0.7\columnwidth]{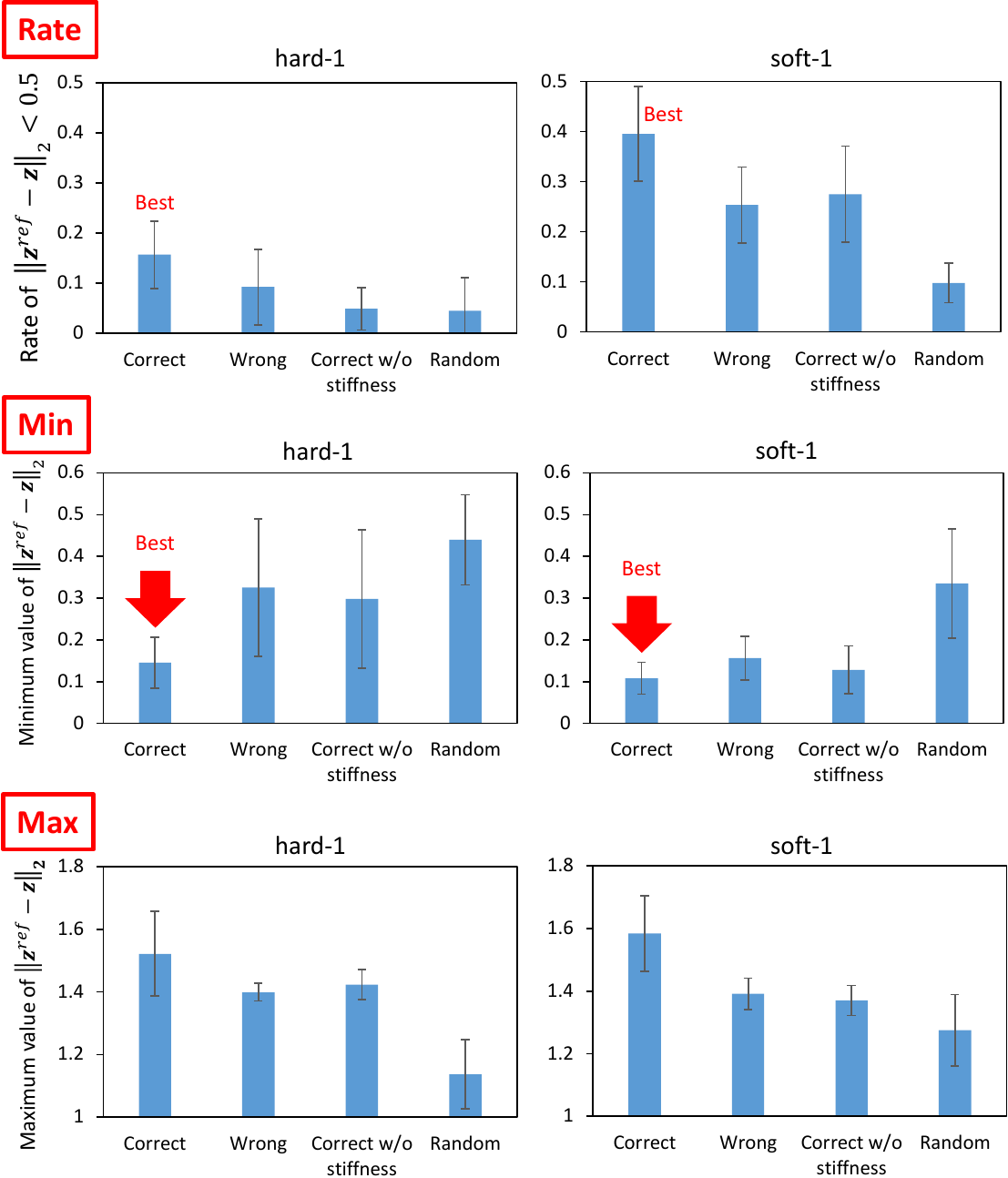}
  \caption{Actual robot experiment: the rate of $||\bm{z}^{ref}-\bm{z}||_{2} < 0.5$ (upper graphs), the minimum value of $||\bm{z}^{ref}-\bm{z}||_{2}$ (middle graphs), and the maximum value of $||\bm{z}^{ref}-\bm{z}||_{2}$ (lower graphs) when conducting experiments of dynamic manipulation of hard-1 and soft-1.}
  \label{figure:eval-all}
\end{figure}

\subsection{Actual Robot Experiment of Musashi-W} \label{subsec:actual-exp}
\switchlanguage%
{%
  Motion data was collected by randomly commanding $\bm{\theta}^{ref}$ and $\bm{k}^{ref}$ (Random) for about 80 seconds, and while human operation with GUI (joint angle is given by GUI, joint stiffness is randomly given) for about 80 seconds, for the cloths of soft-1, soft-2, hard-1, and hard-3, respectively.
  The random command $\bm{k}^{ref}$ is set between [10, 70] [N].
  Using a total of 3000 data points, we trained DPMPB with the number of LSTM expansions being 20, the number of batches being 1000, and the number of epochs being 600 (these parameters are empirically set).
  Principle Component Analysis (PCA) is applied to the parametric bias for each cloth obtained in the training, and its arrangement in the two-dimensional plane is shown in \figref{figure:act-pb}.
  It can be seen that soft-1, soft-2, hard-1, and hard-3 are placed at the four corners, along with the thickness and stiffness of the cloth.

  We execute the control in \secref{subsec:control} and the online material estimation in \secref{subsec:online-estimation} for about 30 seconds with each of the hard-1, hard-2, and hard-3 cloths.
  The trajectory of parametric bias is also shown in \figref{figure:act-pb}.
  Note that the initial value of PB is $\bm{0}$ (the origin of PB is not necessarily at the origin on the figure, since PCA is applied).
  The updated PBs aligned in a straight line according to the thickness of the cloth.
  The updated PB of hard-3 is close to the PB of hard-3 obtained during training, while the updated PB of hard-1 deviated slightly from the PB of hard-1 obtained during training and was close to the PB of hard-3.
  The updated PB of hard-2, which is the data not used in the training, is located midway between hard-1 and hard-3.
  Therefore, we consider that the space of PB is self-organized according to the cloth thickness and cloth stiffness.

  First, we performed the control in \secref{subsec:control} with the value of PB set to the PB obtained during the training of the object to be manipulated.
  From the initial state of the cloth as shown in the left figure of \figref{figure:exp-setup2}, we use \secref{subsec:control} to bring the cloth closer to the state of being spread out in the air as shown in the right figure of \figref{figure:exp-setup2}.
  For hard-1 and soft-1, examples of transitions of the target joint angle $\bm{\theta}^{ref}$, the measured joint angle $\bm{\theta}$, and the muscle tension $f^{const}$ representing the target body stiffness are shown in \figref{figure:test-control}.
  Note that $\theta_{s-p}$ and $\theta_{e-p}$ represent the pitch joint angles of the shoulder and elbow, respectively.
  The area surrounded by the red frame is the main part of the spreading motion, where the upraised shoulder and elbow are lowered down significantly (the larger the joint angle is, the lower the arm is).
  Here, if we look at $f^{const}$, we can see that in many cases, the stiffness is increased in the first half and decreased in the second half.
  This increases the hand speed and causes the cloth to spread out and float in midair, realizing the state shown in the right figure of \figref{figure:exp-setup2}.
  The absolute values of the hand velocities when \equref{eq:backprop} is not performed for $\bm{k}^{ref}$ and $f^{const}$ is fixed ($= 10$ N) (w/o stiffness), and when the control in \secref{subsec:control} is performed (w/ stiffness), are shown in \figref{figure:eval-vel}.
  The figure shows the mean and variance of the maximum hand velocities during five 20-second experiments from the state shown in the left figure of \figref{figure:exp-setup2}.
  It can be seen that the speed is about 12\% faster when the change in the body stiffness is used.
  Also, looking at the difference between soft-1 and hard-1 in \figref{figure:test-control}, in hard-1, $\theta_{e-p}$ and $\theta_{s-p}$ move almost simultaneously and the process from raising to lowering the hand is quick.
  On the other hand, in soft-1, $\theta_{s-p}$ starts to move slightly later than $\theta_{e-p}$, and the process from raising to lowering the hand is often slow.
  The reason is that the soft cloth can maintain flight time even if the hand is lowered slowly at the end, while the hard cloth cannot maintain flight time unless the hand is lowered immediately.
  In this way, we can see that the way of manipulating the object varies depending on the difference in PB.

  Next, we performed the cloth manipulation under several conditions and compared them.
  The trials with correct PBs (the PB of soft-1 is used for the manipulation of soft-1 and the PB of hard-1 is used for the manipulation of hard-1) are called Correct, the trials with wrong PBs (the PB of hard-1 is used for the manipulation of soft-1 and the PB of soft-1 is used for the manipulation of hard-1) are called Wrong, and random trials such as in \secref{subsec:actual-exp} are called Random.
  The case where the PB is correct but \equref{eq:backprop} is not performed for $\bm{k}^{ref}$ and $f^{const}$ is fixed (10N) is called Correct w/o stiffness.
  Under these four conditions, the hard-1 and soft-1 cloths are manipulated for 25 seconds five times from the state shown in the left graph of \figref{figure:exp-setup2}.
  The average of the percentage (y-axis) with $||\bm{z}^{ref}-\bm{z}||_{2}$ lower than a certain threshold (x-axis) is shown in \figref{figure:eval-graph}.
  The means and variances of the proportions of $||\bm{z}^{ref}-\bm{z}||_{2}<0.5$ taken from \figref{figure:eval-graph} are shown in the upper graphs of \figref{figure:eval-all}, the means and variances of the minimum value of $||\bm{z}^{ref}-\bm{z}||_{2}$ are shown in the middle graphs of \figref{figure:eval-all}, and the means and variances of the maximum value of $||\bm{z}^{ref}-\bm{z}||_{2}$ are shown in the lower graphs of \figref{figure:eval-all}.
  \figref{figure:eval-graph} indicates that the more the graph expands in the upper-left direction (the larger the y-axis value becomes when the x-axis value is low), the more the target image is realized.
  Note that the mean of $||\bm{z}^{ref}-\bm{z}||_{2}$ for the initial condition shown in the left figure of \figref{figure:exp-setup2} is 0.86.
  For both hard-1 and soft-1, Correct is the best and Random is the worst.
  And looking at the upper graphs of \figref{figure:eval-all}, we can see the characteristics quantitatively.
  Although hard-1 is more difficult and thus has a lower realization rate in general, the results are consistent with the trend that Wrong and Correct w/o stiffness have lower realization rates than Correct.
  Looking at the middle graphs of \figref{figure:eval-all}, the tendency of the minimum value of $||\bm{z}^{ref}-\bm{z}||_{2}$ is also consistent with the upper graphs of \figref{figure:eval-all}, and the lower the minimum value is, the higher the percentage of realization is.
  On the other hand, for the lower graphs of \figref{figure:eval-all}, Correct is the highest and Random is the lowest.
  This may indicate that the more accurate the method can realize the target image, the more likely it is to produce a state that is different from the target image, since the state must be changed significantly in order to realize the target image.
}%
{%
  soft-1, soft-2, hard-1, hard-3の布をそれぞれ持った状態で, $\bm{\theta}^{ref}$と$\bm{k}^{ref}$をランダムに指令した場合(Random)を約80秒間, GUI (関節角度をGUIで動かし, 関節剛性はランダム)で人間が動作させた場合を約80秒間行いデータを取得した.
  なお, ランダムに指令する$\bm{k}^{ref}$は[10, 70] [N]の間とした.
  計3000のデータを使い, LSTMの展開数を20, バッチ数を1000, エポック数を600としてDPMPBを学習した (これらの値は実験的に決定している, empirically).
  その際に得られたそれぞれの布に関するParametric BiasにPrinciple Component Analysisをかけ, 2次元平面内に配置した様子を\figref{figure:act-pb}に示す.
  soft-1, soft-2, hard-1, hard-3のそれぞれが四隅に配置されており, 布の厚さ・固さ方向が揃っていることがわかる.

  ここで, hard-1, hard-2, hard-3のそれぞれの布を持った状態で, Randomの動作と\secref{subsec:online-estimation}のオンライン素材推定を約30秒間走らせる.
  このときのParametric Biasの軌跡(traj)を\figref{figure:act-pb}に示す.
  なお, PBの初期値は$\bm{0}$とする.
  それぞれのPBは最終的に布の厚みに従って直線上に並んだ.
  hard-3のPBは訓練時に得られたPBに近い値を示す一方, hard-1のPBは訓練時に得られたものから少し外れ, hard-3のPBに寄っていた.
  訓練時に使っていないデータであるhard-2については, hard-1とhard-3の中間程度に位置している.
  よって, PBの空間がCloth thicknessやCloth stiffnessに従って自己組織化されていると考える.

  次に, PBの値を, 操作する物体の訓練時に得られたPBに設定した状態で, \secref{subsec:control}の制御を行った.
  \figref{figure:exp-setup2}の左図のような布の初期状態から, \secref{subsec:control}により, \figref{figure:exp-setup2}の右図にあるような, 布を空中で広げた状態へと近づけていく.
  hard-1とsoft-1をそれぞれ操作したときの指令関節角度$\bm{\theta}^{ref}$, 測定された関節角度$\bm{\theta}$, 指令剛性を表す筋張力$f^{const}$を\figref{figure:test-hard1}, \figref{figure:test-soft1}に示す.
  なお, $\theta_{s-p}$, $\theta_{e-p}$はそれぞれ肩と肘のピッチ関節の角度を表す.
  赤い枠で囲った部分は, 振り上げた肩と肘を大きく下に降ろしている主要な部分である(関節角度が大きいほど下に手を下ろしている状態である).
  このとき, $f^{const}$を見ると, 多くの場合で, 前半は剛性を高くし, 後半は剛性を低くしていることがわかる.
  これにより, 手先スピードが上がり, 布が広がって浮き上がることで\figref{figure:exp-setup2}の右図の状態が実現される.
  実際に, $\bm{k}^{ref}$に対して\equref{eq:backprop}を行わず, $f^{const}$を固定(10N)として実験を行った場合(w/o stiffness)と\secref{subsec:control}を行った場合(w/ stiffness)の手先速度の絶対値を比較した図を\figref{figure:eval-vel}に示す.
  これは, \figref{figure:exp-setup2}の左図の状態から20秒の実験を5回行い, その間の最大手先速度の平均と分散を表したものである.
  剛性変化を使った場合のほうが約12\%程度速度が速くなっていることがわかる.
  また, \figref{figure:test-hard1}, \figref{figure:test-soft1}のsoft-1とhard-1の違いを見ると, hard-1では$\theta_{e-p}$と$\theta_{s-p}$がほぼ同時に動き, また手を振り上げてから下ろすまでが素早い.
  一方, soft-1では, $\theta_{s-p}$が$\theta_{e-p}$よりも少し後に動き出しており, また手を振り上げてから振り下ろすまでが緩やかな場合が多い.
  これは, 柔らかい布は最後はゆったりと手をおろしても滞空時間が確保できるのに対して, 固い布はすぐに手を降ろさないと滞空時間が確保できないことが理由であると考えられる.
  このように, PBの違いによって対象の操作の仕方が変化することがわかる.

  最後に, いくつかの条件で布操作を行い, それらを比較した.
  PBが正しい(soft-1の操作に対してsoft-1, hard-1の操作に対してhard-1のPBを用いている)場合の試行をCorrect, PBが間違っている(soft-1の操作に対してhard-1, hard-1の操作に対してsoft-1のPBを用いている)場合の試行をWrong, ランダムな試行をRandomとする.
  また, PBは正しいが, $\bm{k}^{ref}$に対して\equref{eq:backprop}を行わず, $f^{const}$を固定(10N)として実験を行った場合をCorrect w/o stiffnessとする.
  この4つの条件において, hard-1とsoft-1のそれぞれの布を\figref{figure:exp-setup2}の左図の状態から25秒の実験を5回行う.
  その際に, $||\bm{z}^{ref}-\bm{z}||_{2}$がある閾値(x軸)よりも低い割合(y軸)の平均を\figref{figure:eval-graph}に示す.
  また, \figref{figure:eval-graph}の一部を取り出した, $||\bm{z}^{ref}-\bm{z}||_{2}<0.5$の割合の平均と分散を\figref{figure:eval-sum}に, $||\bm{z}^{ref}-\bm{z}||_{2}$の最小値の平均と分散を\figref{figure:eval-min}に, $||\bm{z}^{ref}-\bm{z}||_{2}$の最大値の平均と分散を\figref{figure:eval-max}に示す.
  なお, \figref{figure:exp-setup2}の左図の初期状態における$||\bm{z}^{ref}-\bm{z}||_{2}$の平均は0.86であった.
  まず, \figref{figure:eval-graph}はグラフが左上方向に膨らむほど(xの値が小さいときにyの割合が多いほど), より指令画像状態が実現できていることを表す.
  hard-1, soft-1のどちらにおいてもCorrectが最も良く, Randomが最も悪い.
  そして, \figref{figure:eval-sum}を見ると, その特性が定量的に見て分かる.
  hard-1の方がより難しいため全体的に実現率が低くいが, Correctに対してWrong, Correct w/o stiffnessの実現率が低いという傾向は一致している.
  \figref{figure:eval-min}を見ると, $||\bm{z}^{ref}-\bm{z}||_{2}$の最小値の傾向も\figref{figure:eval-sum}と一致しており, 最小値が低いものほど実現率が高い割合も大きい.
  一方, \figref{figure:eval-max}はCorrectが最も高く, Randomが最も低い.
  これは, 指令画像を実現するためには大きく動的に状態を変化させなければならないため, より指令画像を正しく実現できる方法ほど指令画像と離れた状態が起きやすいことを表していると考えられる.
}%

\begin{figure}[t]
  \centering
  \includegraphics[width=1.0\columnwidth]{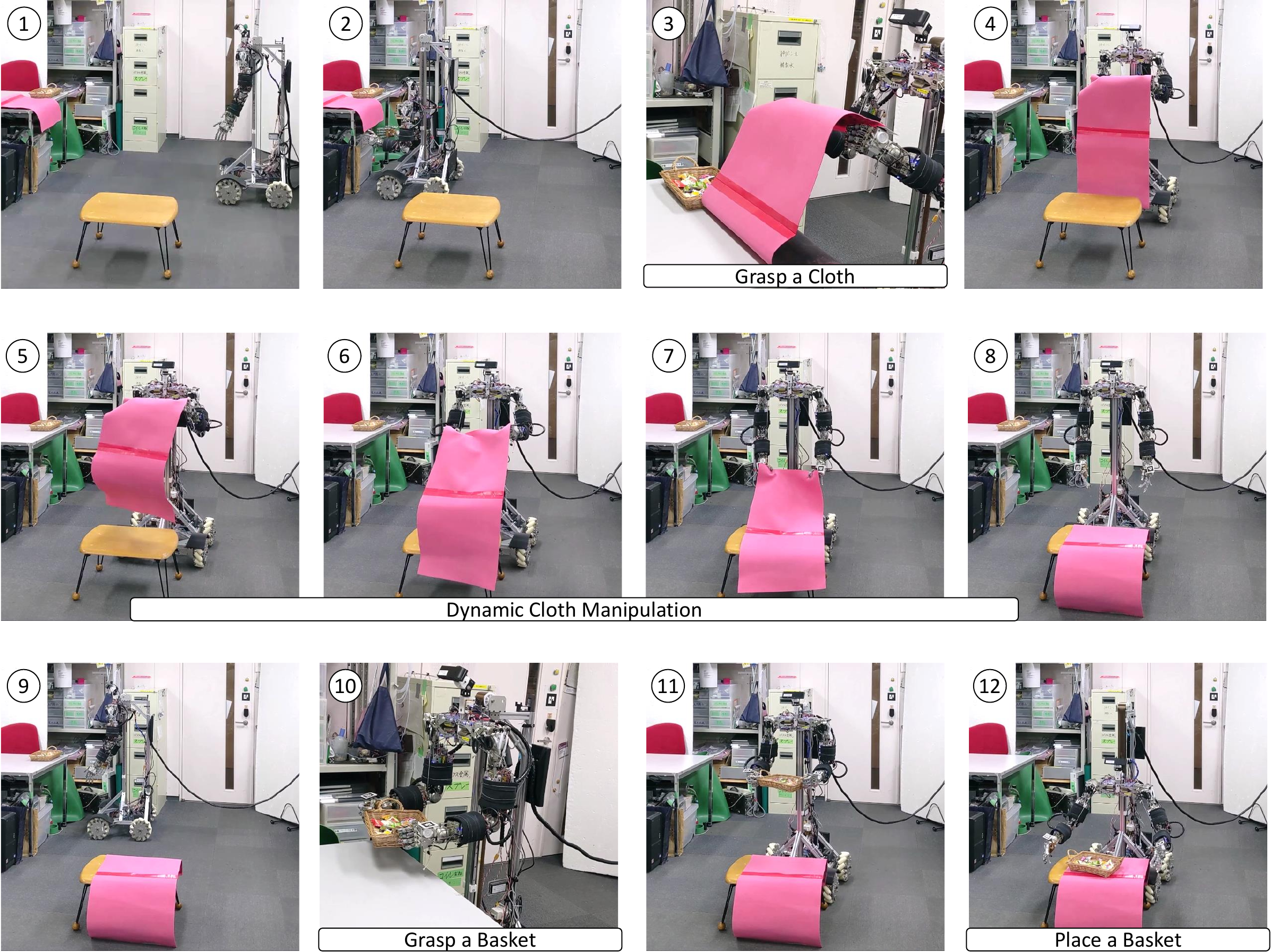}
  \caption{The integrated table setting experiment of grasping a cloth, manipulating it dynamically, and putting a basket on it.}
  \label{figure:integrated-exp}
\end{figure}

\subsection{Integrated Table Setting Experiment} \label{subsec:integrated-exp}
\switchlanguage%
{%
  We performed a series of motion experiments incorporating dynamic cloth manipulation.
  Musashi-W picked up hard-1 and laid it on the table like a table cloth by the proposed method, and placed a basket of sweets on it.
  The scene is shown in \figref{figure:integrated-exp}.
  First, the robot recognizes the point cloud of the cloth and grasps the cloth between its thumb and index by visual feedback.
  Next, the robot spreads the cloth on the desk by dynamic cloth manipulation, which has been previously trained on a similar setup.
  Finally, the robot recognizes the point cloud of the basket and grasps the basket between both hands by visual feedback, and successfully places it on the table.
}%
{%
  動的布操作を含む一連の動作実験を行った.
  Musashi-Wがhard-1を手に取り, 本手法によりテーブルの上にテーブルクロスとしてこれを敷き, その上にお菓子の入ったかごを載せる.
  その様子を\figref{figure:integrated-exp}に示す.
  まず, 布のポイントクラウドを認識してビジュアルフィードバックにより親指と人差し指の間で布を挟み, これを把持する.
  次に, 予め同様の机等のセットアップで学習された動的布操作により, この布を机の上に広げる.
  最後に, カゴのポイントクラウドを認識してビジュアルフィードバックによりかごを両手で挟んで把持し, これをテーブルの上に置くことに成功した.
}%

\section{Discussion} \label{sec:discussion}
\switchlanguage%
{%
  We discuss the results obtained from the experiments in this study.
  First, simulation experiments show that the value of PB is self-organized according to the parameters of the cloth.
  The dynamics of the current cloth can be estimated online from the cloth manipulation data.
  On the other hand, we found that it is difficult to accurately estimate the parameters that do not contribute much to the change in dynamics.
  Through dynamic cloth manipulation experiments while changing the cloth parameters and the target image, our method can accurately realize the target image.
  By changing PB, the method can handle various cloth properties, and the target image can be given arbitrarily.
  The reproducibility of the performance on learning and control is also high.
  From the integrated experiment of online material estimation and dynamic cloth manipulation, we show that they can be performed simultaneously, and that the closer the current PB is to the current cloth properties, the more accurately the target image can be realized.

  From the actual robot experiments, we found that the value of PB is self-organized depending on the thickness and stiffness of cloths.
  As in the simulation experiments, the dynamics of the current cloth can be estimated online from the cloth manipulation data.
  We also found that the dynamics can be estimated at the correct point, which is the internally dividing point of the trained PBs, even for the data that is not available at the time of training.
  Through dynamic cloth manipulation experiments, we found that the behavior of the cloth manipulation varies depending on the PB which expresses the dynamics of the cloth.
  The joint angles and the speed were appropriately changed according to the material of the cloth to be manipulated.
  In addition, the stiffness value affects the speed of the hand movements, and when the stiffness value is optimized, the hand speed is increased by about 12\% compared to the case without its optimization.
  When PB does not match the current cloth or the stiffness value is not optimized, the performance decreased compared to the case where PB is correct and the variable stiffness is used.
  In addition, we found that in order to realize the target image correctly, it is necessary to go through a state that is far from the target image.
  From these results, we found that the dynamics of the cloth material is embedded and estimated online through parametric bias, that variable stiffness control can be used explicitly to improve the hand speed, and that the target cloth image can be realized more accurately by using deep predictive model learning and backpropagation technique to the control command.

  While our method can greatly improve the ability of dynamic cloth manipulation of robots, there are still some issues to be solved.
  First, in this study, the body stiffness is substituted by a certain single value, but in humans, the body stiffness can be set more flexibly.
  It is necessary to discuss the degree of freedom of the body stiffness in the future.
  Next, we consider that it is difficult for our method to deal with the case of large nonlinear changes such as the cloth leaving the hand or changing the grasping position.
  There is no method that can handle such highly nonlinear, discontinuous, and high-dimensional states in trials using only actual robots, and this is an issue for the future.
  Finally, we are still far from reaching human-like adaptive dynamic cloth manipulation.
  Humans are capable of stretching and spreading the cloth by focusing on small wrinkles, instead of just looking at the entire cloth universally.
  In addition, since the current hardware is not as agile and soft as humans, it will be necessary to develop both hardware and software.
}%
{%
  本研究の実験から得られた結果について議論する.
  まず, シミュレーション実験から, PBの値が布のパラメータに応じて自己組織化されることがわかった.
  布の操作動作データから, 現在持っている布のダイナミクスをオンラインで推定することができる.
  一方で, ダイナミクスにあまり寄与しないパラメータについては, 正確に推定することが難しいこともわかった.
  次に, 布の特性, 指令画像を変化させながら動的布操作実験を行うことで, 本手法が正確に指令画像を実現できることがわかった.
  PBを変化させることで, 様々な布の特性に対応でき, また, 指令画像も任意に与えることが可能である.
  また, 学習や制御に関する性能の再現性も高い.
  最後に, オンライン素材推定と動的布操作の統合実験から, これらは同時に実行できること, また, 現在のPBが現在の布の特性に近くなるほどより正確に指令画像を実現できるようになることがわかった.

  実機実験から, PBの値が布の固さ・枚数に応じて自己組織化されることがわかった.
  シミュレーション同様, 布の操作動作データから, 現在持っている布のダイナミクスをオンラインで推定することができる.
  また, 訓練時には無いデータについても訓練時のデータを内分するような正しい点にダイナミクスが推定されることがわかった.
  次に, PB, つまり布のダイナミクスに応じて布操作の動作が大きく変化することがわかった.
  関節角度の連動のさせ方やスピードなどが, 操作する布の素材に合わせて適切に変更されることがわかった.
  また, 剛性値は手先の動作スピードに影響を及ぼし, これを最適化した場合は, 最適化しない場合に比べて手先速度が12\%程度上昇することがわかった.
  PBが現在の布に合致していない場合や剛性値を最適化しない場合は, PBが正しく, 可変剛性を利用する場合に比べると大きく性能が低下した.
  また, 指令画像を正しく実現するためには, 指令画像から大きく離れた状態を経由する必要があることがわかった.
  これらから, Parametric Biasによって布素材のダイナミクスの埋め込み・オンラインの推定ができること, 可変剛性制御を陽に利用することで手先速度を向上させることができること, これらを統合した深層予測モデル学習と制御入力への誤差逆伝播を用いることで, より正確に指令布状態を実現できることがわかった.

  本手法はロボットの動的布操作の能力を大きく底上げすることができる一方, まだ課題も残る.
  まず, 本研究では身体剛性をある一定値により代替しているが, 実際の人間ではより細かく身体剛性を設定することができる.
  今後, 身体剛性の自由度についても議論する必要がある.
  次に, 本研究は手から布が離れたり, 布を持ち替えたり等, 大きく非線形な変化が起こる場合に対処することは難しいと考える.
  実機のみでの試行では, そのような高い非線形性・不連続性・高い次元の状態を扱うことができる手法はなく, 今後の課題である.
  最後に, まだ人間の動的布操作への到達にはほど遠い.
  人間は, 柔軟物体全体を万遍なく見るのではなく, 細かいシワに着目しながらそれを伸ばしつつ広げることも可能である.
  また, 現状人間ほど俊敏に柔らかく動けるハードウェアではないため, 今後ハードウェア・ソフトウェアの両面からの開発が必要となるだろう.
}%

\section{Conclusion} \label{sec:conclusion}
\switchlanguage%
{%
  In this study, we developed a deep predictive model learning method that incorporates quick manipulation using variable stiffness mechanism and response to changes in cloth material for more human-like dynamic cloth manipulation.
  For variable stiffness, the command value is calculated by backpropagation technique using the joint angle and the body stiffness value as control commands.
  For the adaptation to change in cloth material, we embed information about the cloth dynamics into parametric bias.
  The dynamics of the cloth material can be estimated online even when it is not included in the training data, and it is shown that the characteristics of the dynamic cloth manipulation varies greatly depending on the material.
  In addition, when the stiffness value is appropriately set according to the motion phase, the speed can be increased by about 12\% compared to the case without variable stiffness control, and the target cloth state can be realized more accurately.
  In the future, we would like to develop a method to handle more complicated cloth manipulation tasks in a unified manner.
}%
{%
  本研究では, より人間らしい動的布操作を目指し, 布の素材変化への対応・可変剛性を使った素早いマニピュレーションを取り入れた深層予測モデル学習手法を開発した.
  布の素材変化に対してはParametric Biasを, 可変剛性については関節角度と剛性値を制御入力として誤差逆伝播を用いて指令値を計算する手法を取り入れた.
  訓練データにはない布素材についてもそのダイナミクスが推定でき, また, 素材によって動的布操作の特性が大きく変わることを示した.
  また, 動作フェーズに応じて適切に剛性値が設定されることで, 可変剛性制御を行わない場合に比べて速度を12\%程度向上させ, より正確に指令状態を実現することが可能であった.
  今後, より複雑な布操作タスクも統一的に扱えるような手法へと発展させていきたい.
}%

\bibliographystyle{include/frontiersinSCNS_ENG_HUMS} % for Science, Engineering and Humanities and Social Sciences articles, for Humanities and Social Sciences articles please include page numbers in the in-text citations
\bibliography{main}

\end{document}